# A multilevel thresholding algorithm using Electro-magnetism Optimization

Diego Oliva [a], [1]Erik Cuevas[b], Gonzalo Pajares [a], Daniel Zaldivar [b], Valentín Osuna [c]

[a] Dpt. Ingeniería del Software e Inteligencia Artificial,
Facultad Informática, Universidad Complutense,
28040 Madrid, Spain
doliva@estumail.ucm.es
pajares@fdi.ucm.es

[b] Departamento de Ciencias Computacionales
Universidad de Guadalajara, CUCEI
Av. Revolución 1500, Guadalajara, Jal, México
{[1]erik.cuevas, daniel.zaldivar }@cucei.udg.mx

[b] Centro de Investigación en Computación-IPN
Av. Juan de Dios Batiz S/N, Mexico, D. F. MEXICO
josuna_b11@sagitario.cic.ipn.mx

**Abstract**

Segmentation is one of the most important tasks in image processing. It consist in classify the pixels into two or more groups depending on their intensity levels and a threshold value. The quality of the segmentation depends on the method applied to select the threshold. The use of the classical implementations for multilevel thresholding is computationally expensive since they exhaustively search the best values to optimize the objective function. Under such conditions, the use of optimization evolutionary approaches has been extended. The Electro-magnetism-Like algorithm (EMO) is an evolutionary method which mimics the attraction-repulsion mechanism among charges to evolve the members of a population. Different to other algorithms, EMO exhibits interesting search capabilities whereas maintains a low computational overhead. In this paper, a multilevel thresholding (MT) algorithm based on the EMO is introduced. The approach combines the good search capabilities of EMO algorithm with objective functions proposed by the popular MT methods of Otsu and Kapur. The algorithm takes random samples from a feasible search space inside the image histogram. Such samples build each particle in the EMO context whereas its quality is evaluated considering the objective that is function employed by the Otsu's or Kapur's method. Guided by these objective values the set of candidate solutions are evolved through the EMO operators until an optimal solution is found. The approach generates a multilevel segmentation algorithm which can effectively identify the threshold values of a digital image in a reduced number of iterations. Experimental results show performance evidence of the implementation of EMO for digital image segmentation.

## 1. Introduction

Image processing has several applications in areas as medicine, industry, agriculture, etc. Most of all the methods of image processing require a first step called segmentation. This task consists in classify the pixels in the image depending on its gray (or RGB in each component) level intensity (histogram). In this way, several techniques had been studied [1, 10]. Thresholding is the easiest method for segmentation as it works taking a threshold ( *th* ) value and the pixels which intensity value is higher than *th* are labeled as the first class and the rest of the pixels correspond to a second class. When the image is segmented into two classes, the task is called bi-level thresholding (BT) and it requires only one *th* value. On the other hand, when pixels are separated into more than two classes, the task is named as multilevel thresholding (MT) and demands more than one *th* values [2, 10]. Threshold based methods are divided into parametric and nonparametric [2-4]. For parametric approaches it is necessary to estimate some parameters of a probability density function which models each class. Such approaches are time consuming and computationally expensive. On the other hand, the nonparametric employs several criteria such as between-class variance, the entropy and the error rate [5-7] that must be optimized to determine the optimal threshold values. These approaches result an attractive option due their robustness and accuracy [8].

---
[1] Corresponding author, Tel +52 33 1378 5900, ext. 7715, E-mail: erik.cuevas@cucei.udg.mx

For bi-level thresholding there exist two classical methods, the first maximizes the between classes variance and was proposed by Otsu [5]. The second submitted by Kapur in [6] uses the maximization of the entropy to measure the homogeneity of the classes. Their efficiency and accuracy have been already proved for a bi-level segmentation[9]. Although both Otsu's and Kapur's can be expanded for multilevel thresholding, their computational complexity increases exponentially with each new threshold [9].

As an alternative to classical methods, the MT problem has also been handled through evolutionary optimization methods. In general, they have demonstrated to deliver better results than those based on the classical techniques in terms of accuracy, speed and robustness. Numerous evolutionary approaches have been reported in the literature. Hammouche et al. provides a survey of different evolutionary algorithms such as (Differential Evolution (DE), Simulated Annealing (SA), Tabu Search (TS) etc.), used to solve the Kaptur's and Otsu's problems [2]. In [2,11,12], Genetic Algorithms-based approaches are employed to segment multi-classes. Similarly in [1,5], Particle Swarm Optimization (PSO) [13] has been proposed for MT proposes, maximizing the Otsu's function. Other examples such as [14-16] including Artificial Bee Colony (ABC) or Bacterial Foraging Algorithm (BFA) for image segmentation.

This paper introduces a multilevel threshold method based on the Electromagnetism-like Algorithm (EMO). EMO is a global optimization algorithm that mimics the electromagnetism law of physics. It is a population-based method which has an attraction-repulsion mechanism to evolve the members of the population guided by their objective function values [17]. The main idea of EMO is to move a particle through the space following the force exerted by the rest of the population. The force is calculated using the charge of each particle based on its objective function value. Unlike other meta-heuristics such as GA, DE, ABC and Artificial Immune System (AIS), where the population members exchange materials or information between each other, in EMO similar to heuristics such as PSO and Ant Colony Optimization (ACO) each particle is influenced by all other particles within its population. Although the EMO algorithm shares some characteristics to PSO and ACO, recent works have exhibited its better accuracy regarding optimal parameters [18 - 21], yet showing convergence [22]. In recent works, EMO has been used to solve different sorts of engineering problems such as flow-shop scheduling [23], communications [24], vehicle routing [25], array pattern optimization in circuits [26], neural network training [27], image processing [28] and control systems [29]. Although EMO algorithm shares several characteristics to other evolutionary approaches, recent works (see [18-21]) have exhibited a better EMO's performance in terms of computation time and precision when it is compared with other methods such as GA, PSO and ACO.

In this paper, a segmentation method called Multilevel Threshold based on the EMO algorithm (MTEMO) is introduced. The algorithm takes random samples from a feasible search space which depends on the image histogram. Such samples build each particle in the EMO context. The quality of each particle is evaluated considering the objective function employed by the Otsu's or Kapur's method. Guided by this objective value the set of candidate solutions are evolved using the attraction-repulsion operators. The approach generates a multilevel segmentation algorithm which can effectively identify the threshold values of a digital image within a reduced number of iterations and decreasing the computational complexity of the original proposals. Experimental results show performance evidence of the implementation of EMO for digital image segmentation.

The rest of the paper is organized as follows. In Section 2, the standard EMO algorithm is introduced. Section 3 gives a simple description of the Otsu's and Kapur's methods. Section 4 explains the implementation of the proposed algorithm. Section 5 discusses experimental results and comparisons after test the MTEMO in a set benchmark images. Finally, the work is concluded in Section 6.

**2. Electromagnetism – Like Optimization Algorithm (EMO)**

The EMO method has been designed to solve the problem of finding a global solution of a nonlinear optimization problem with box constraints in the following form:

$$\begin{aligned} &\text{maximize} & &f(x), \quad x=(x_1,\ldots,x_n) \in \Re^n \\ &\text{subject to} & &x \in \mathbf{X} \end{aligned} \quad (1)$$

where $f: \Re^n \to \Re$ is a nonlinear function whereas $\mathbf{X} = \{x \in \Re^n | l_i \leq x_i \leq u_i, i = 1,\ldots,n\}$ is a bounded feasible region, constrained by the lower ($l_i$) and upper ($u_i$) limits.

EMO [17] utilizes $N$, $n$-dimensional points $x_{i,t}$, as a population for exploring the feasible set $\mathbf{X}$, where $t$ denotes the iteration (or generation) number of the algorithm. The initial population $\mathbf{S}_t = \{x_{1,t}, x_{2,t}, \ldots, x_{N,t}\}$ (being $t=1$), is taken of uniformly distributed samples of the search region, $\mathbf{X}$. We denote the population set at the $t$-th iteration by $\mathbf{S}_t$, as the members of $\mathbf{S}_t$ changes with $t$. After the initialization of $\mathbf{S}_t$, EMO continues its iterative process until a stopping condition (e.g. the maximum number of iterations) is met. An iteration of EMO consists of two steps. In the first step, each point in $\mathbf{S}_t$ moves to a different location by using the attraction-repulsion mechanism of the electromagnetism theory [30]. In the second step, points moved by the electromagnetism principle are further moved locally by a local search and then become members of $\mathbf{S}_{t+1}$ in the $(t+1)$-th iteration. Both the attraction-repulsion mechanism and the local search in EMO are responsible for driving the me members, $x_{i,t}$, of $\mathbf{S}_t$ to the close proximity of the global optimizer.

As with the electromagnetism theory for charged particles, each point $x_{i,t} \in \mathbf{S}_t$ in the search space $\mathbf{X}$ is assumed as a charged particle where the charge of a point relates to its objective function value. Points with better objective function value have more charges than other points, and the attraction-repulsion mechanism is a process in EMO by which points with more charge attract other points in $\mathbf{S}_t$, and points with less charge repel other points. Finally, a total force vector $F_i^t$, exerted on a point e.g. the $i$-th point $x_{i,t}$ is calculated by adding these attraction – repulsion forces and each $x_{i,t} \in \mathbf{S}_t$ is moved in the direction of its total force to the location $y_{i,t}$. A local search is used to explore the vicinity of the each $y_{i,t}$ by $y_{i,t}$ to $z_{i,t}$. The members, $x_{i,t+1} \in \mathbf{S}_{t+1}$, of the $(t+1)$-th iteration are then found by using:

$$x_{i,t+1} = \begin{cases} y_{i,t} & \text{if } f(y_{i,t}) \leq f(z_{i,t}) \\ z_{i,t} & \text{otherwise} \end{cases} \qquad (2)$$

Algorithm 1 shows the general scheme of EMO. We also provided the description of each step following the algorithm.

---

**Algorithm 1** [EMO ($N, Iter_{max}, Iter_{local}, \delta$)]

1. Input parameters: Maximum number of iteration $Iter_{max}$, values for the local search parameter such $Iter_{local}$ and $\delta$, and the size $N$ of the population.
2. Initialize: set the iteration counter $t=1$, initialize the number of $\mathbf{S}_t$ uniformly in $\mathbf{X}$ and identify the best point in $\mathbf{S}_t$.
3. while $t < Iter_{max}$ do
4.      $F_i^t \leftarrow \text{CalcF}(\mathbf{S}_t)$
5.      $y_{i,t} \leftarrow \text{Move}(x_{i,t}, F_i^t)$
6.      $z_{i,t} \leftarrow \text{Local}(Iter_{local}, \delta, y_{i,t})$
7.      $x_{i,t+1} \leftarrow \text{Select}(\mathbf{S}_{t+1}, y_{i,t}, z_{i,t})$
8. end while

---

Input parameters (Line 1): EMO algorithm runs for $Iter_{max}$ iterations. In the local search phase, $n \times Iter_{local}$ is the maximum number of locations $z_{i,t}$, within a $\delta$ distance of $y_{i,t}$, for each $i$ dimension.

Initialize (Line 2): The points $x_{i,t}$, $t=1$, are selected uniformly in $\mathbf{X}$, i.e. $x_{i,1} \sim Unif(\mathbf{X})$, $i=1,2,...,N$, where *Unif* represents the uniform distribution. The objective function values $f(x_{i,t})$ are computed, and the best point is identified as follows:

$$x_t^B = \arg \max_{x_{i,t} \in \mathbf{S}_t} \{f(x_{i,t})\}, \qquad (3)$$

where $x_t^B$ is the element of $\mathbf{S}_t$ that produces the maximum value in terms of the objective function $f$.

Calculate force (Line 4): In this step, a charged-like value ($q_{i,t}$) is assigned to each point ($x_{i,t}$). The charge $q_{i,t}$ of $x_{i,t}$ depends on $f(x_{i,t})$ and points with better objective function have more charge than others. The charges are computed as follows:

$$q_{i,t} = \exp\left(-n \frac{f(x_{i,t}) - f(x_t^B)}{\sum_{j=1}^{N} f(x_{i,t}) - f(x_t^B)}\right) \qquad (4)$$

Then the force, $F_{i,j}^t$, between two points $x_{i,t}$ and $x_{j,t}$ is calculated using:

$$F_{i,j}^t = \begin{cases} (x_{j,t} - x_{i,t}) \dfrac{q_{i,t} \cdot q_{j,t}}{\|x_{j,t} - x_{i,t}\|^2} & \text{if } f(x_{i,t}) > f(x_{j,t}) \\ (x_{i,t} - x_{j,t}) \dfrac{q_{i,t} \cdot q_{j,t}}{\|x_{j,t} - x_{i,t}\|^2} & \text{if } f(x_{i,t}) \le f(x_{j,t}) \end{cases} \qquad (5)$$

The total force, $F_i^t$, corresponding to $x_{i,t}$ is now calculated as:

$$F_i^t = \sum_{j=1, j \ne i}^{N} F_{i,j}^t \qquad (6)$$

Move the point $x_{i,t}$ along $F_i^t$ (Line 5): In this step, each point $x_{i,t}$ except for $x_t^B$ is moved along the total force $F_i^t$ using:

$$x_{i,t} = x_{i,t} + \lambda \frac{F_i^t}{\|F_i^t\|}(RNG), \quad i=1,2,...,N; \ i \ne B \qquad (7)$$

where $\lambda \sim Unif(0,1)$ for each coordinate of $x_{i,t}$, and *RNG* denotes the allowed range of movement toward the lower or upper bound for the corresponding dimension.

Local search (Line 6): For each $y_{i,t}$ a maximum of $iter_{local}$ points are generated in each coordinate direction in the $\delta$ neighbourhood of $y_{i,t}$. This means that the process of generating local point is continued for each $y_{i,t}$ until either a better $z_{i,t}$ is found or the $n \times Iter_{local}$ trial is reached.

Selection for the next iteration (Line 7): In this step, $x_{i,t+1} \in \mathbf{S}_{t+1}$ are selected from $y_{i,t}$ and $z_{i,t}$ using Eq (1), and the best point is identified by using Eq. (3).

All evolutionary methods have been designed in the way that regardless of the starting point, there exists a good probability to find either the global optima or a good enough sub-optimal solution. However, most of approaches lack of a formal proof of such convergence. One exception is the EMO algorithm for which a complete convergence analysis has been developed in [22]. Such study assumes a bound-constrained optimization problem and demonstrates the existence of a considerable probability of at least one particle of the population $\mathbf{S}_t$ moving closer to the set of optimal solutions after only one iteration. Therefore, the EMO method can effectively deliver the solution for complex optimization problems yet requiring a low number of iterations in comparison to other evolutionary methods. Such a fact has been demonstrated through several experimental studies for EMO [25,27,31,32] where its computational cost and its iteration number have been compared to other evolutionary methods for the case of several engineering related problems.

## 3. Image Multilevel Thresholding (MT)

Thresholding is a process in which the pixels of a gray scale image are divided in sets or classes depending on their intensity level ($L$). For this classification it is necessary to select a threshold value ($th$) and follows the simple rule of Eq. (8).

$$C_1 \leftarrow p \quad \text{if} \quad 0 \leq p < th$$
$$C_2 \leftarrow p \quad \text{if} \quad th \leq p < L-1 \tag{8}$$

Where $p$ is one of the $m \times n$ pixels of the gray scale image $I_g$ that can be represented in $L$ gray scale levels $L = \{0,1,2,...,L-1\}$. $C_1$ and $C_2$ are the classes in which the pixel $p$ can be located, while $th$ is the threshold. The rule in Ec. (8) corresponds to a bi-level thresholding and can be easily extended for multiple sets:

$$C_1 \leftarrow p \quad \text{if} \quad 0 \leq p < th_1$$
$$C_2 \leftarrow p \quad \text{if} \quad th_1 \leq p < th_2$$
$$C_i \leftarrow p \quad \text{if} \quad th_i \leq p < th_{i+1}$$
$$C_n \leftarrow p \quad \text{if} \quad th_n \leq p < L-1 \tag{9}$$

where $\{th_1 \ th_2 \ ... \ th_i \ th_{i+1} \ th_k\}$ represent the different thresholds. The problem for both bi-level and multilevel thresholding is to select the $th$ values that correctly identify the classes. Otsu's and Kapur's methods are well-known approaches for determining such values. Both methods propose a different objective function which must be maximized in order to find optimal threshold values, just as it is discussed below.

*3.1 Between – class variance (Otsu's method)*

This is a nonparametric technique for thresholding proposed by Otsu [5] that employs the maximum variance value of the different classes as a criterion to segment the image. Taking the $L$ intensity levels from an intensity image or from each component of a RGB (red, green, blue) image, the probability distribution of the intensity values is computed as follows:

$$Ph_i^c = \frac{h_i^c}{NP}, \quad \sum_{i=1}^{NP} Ph_i^c = 1, \quad c = \begin{cases} 1,2,3 & \text{if} \quad \text{RGB Image} \\ 1 & \text{if} \quad \text{Gray scale Image} \end{cases} \tag{10}$$

where $i$ is a specific intensity level ($0 \leq i \leq L-1$), $c$ is the component of the image which depends if the image is intensity or RGB whereas $NP$ is the total number of pixels in the image. $h_i^c$ (histogram) is the number of pixels that corresponds to the $i$ intensity level in $c$. The histogram is normalized in a probability distribution $Ph_i^c$. For the simplest segmentation (bi-level) two classes are defined as:

$$C_1 = \frac{Ph_1^c}{\omega_0^c(th)}, \ldots, \frac{Ph_{th}^c}{\omega_0^c(th)} \quad \text{and} \quad C_2 = \frac{Ph_{th+1}^c}{\omega_1^c(th)}, \ldots, \frac{Ph_L^c}{\omega_1^c(th)} \tag{11}$$

where $\omega_0(th)$ and $\omega_1(th)$ are probabilities distributions for $C_1$ and $C_2$, as it is shown by Eq. (12).

$$\omega_0^c(th) = \sum_{i=1}^{th} Ph_i^c, \quad \omega_1^c(th) = \sum_{i=th+1}^{L} Ph_i^c \tag{12}$$

It is necessary to compute the mean levels $\mu_0^c$ and $\mu_1^c$ that define the classes using Eq. (13). Once those values are calculated, the Otsu based between – class $\sigma_B^{2c}$ is calculated using Eq. (14).

$$\mu_0^c = \sum_{i=1}^{th} \frac{iPh_i^c}{\omega_0^c(th)}, \quad \mu_1^c = \sum_{i=th+1}^{L} \frac{iPh_i^c}{\omega_1^c(th)} \tag{13}$$

$$\sigma_B^{2c} = \sigma_1^c + \sigma_2^c \tag{14}$$

Notice that for both Equations Eq. (13) and Eq. (14), $c$ depends on the type of image. Moreover $\sigma_1^c$ and $\sigma_2^c$ in Eq. (14) are the variances of $C_1$ and $C_2$ which are defined as:

$$\sigma_1^c = \omega_0^c \left( \mu_0^c + \mu_T^c \right)^2, \quad \sigma_2^c = \omega_1^c \left( \mu_1^c + \mu_T^c \right)^2 \tag{15}$$

where $\mu_T^c = \omega_0^c \mu_0^c + \omega_1^c \mu_1^c$ and $\omega_0^c + \omega_1^c = 1$. Based on the values $\sigma_1^c$ and $\sigma_2^c$, Eq. (16) presents the objective function. Therefore, the optimization problem is reduced to find the intensity level that maximizes Eq. (16).

$$f_{Otsu}(th) = \max(\sigma_B^{2c}(th)), \quad 0 \leq th \leq L-1 \tag{16}$$

Where $\sigma_B^{2c}(th)$ is the Otsu´s variance for a given $th$ value. Therefore, the optimization problem is reduced to find the intensity levels ($th$) that maximizes Eq. (16).

Otsu's method is applied for a single component of an image, what means for RGB images it is necessary to apply separation into single component images. The previous description of such bi-level method can be extended for the identification of multiple thresholds. Considering $k$ thresholds it is possible separate the original image into $k$ classes using Eq. (9), then it is necessary to compute the $k$ variances and their respective elements. The objective function $f_{Otsu}(th)$ in Eq. (16) can thus be rewritten for multiple thresholds as follows:

$$f_{Otsu}(\mathbf{TH}) = \max(\sigma_B^{2c}(\mathbf{TH})), \quad 0 \leq th_i \leq L-1, \quad i = 1, 2, \ldots, k \tag{17}$$

where $\mathbf{TH} = [th_1, th_2, ..., th_{k-1}]$, is a vector containing multiple thresholds and the variances are computed through Eq. (18).

$$\sigma_B^{2^c} = \sum_{i=1}^{k} \sigma_i^c = \sum_{i=1}^{k} \omega_i^c \left(\mu_i^c - \mu_T^c\right)^2 \tag{18}$$

Here $i$ represents and specific class. $\omega_i^c$ and $\mu_j^c$ are respectively the probability of occurrence and the mean of a class, respectively. For MT such values are obtained as:

$$\omega_0^c(th) = \sum_{i=1}^{th_1} Ph_i^c \tag{19}$$

$$\omega_1^c(th) = \sum_{i=th_1+1}^{th_2} Ph_i^c$$

$$\vdots \qquad \vdots$$

$$\omega_{k-1}^c(th) = \sum_{i=th_k+1}^{L} Ph_i^c$$

and for the mean values:

$$\mu_0^c = \sum_{i=1}^{th_1} \frac{iPh_i^c}{\omega_0^c(th_1)} \tag{20}$$

$$\mu_1^c = \sum_{i=th_1+1}^{th_2} \frac{iPh_i^c}{\omega_0^c(th_2)}$$

$$\vdots \qquad \vdots$$

$$\mu_{k-1}^c = \sum_{i=th_k+1}^{L} \frac{iPh_i^c}{\omega_1^c(th_k)}$$

Similar to the bi-level case, for the MT using the Otsu's method $c$ corresponds to the image components, RGB $c = 1, 2, 3$ and intensity $c = 1$.

*3.2 Entropy criterion method (Kapur's method)*

Another nonparametric method that is used to determine the optimal threshold values has been proposed by Kapur [6]. It is based on the entropy and the probability distribution of the image histogram. The method aims to find the optimal $th$ that maximizes the overall entropy. The entropy of an image measures the compactness and separability among classes. In this sense when the optimal $th$ value appropriately separates the classes, the entropy has the maximum value. For the bi- level example the objective function of the Kapur's problem can be defined as:

$$f_{Kapur}(th) = H_1^c + H_2^c, \quad c = \begin{cases} 1, 2, 3 & \text{if} \quad \text{RGB Image} \\ 1 & \text{if} \quad \text{Gray scale Image} \end{cases} \tag{21}$$

where the entropies $H_1$ and $H_2$ are computed by the following model:

$$H_1^c = \sum_{i=1}^{th} \frac{Ph_i^c}{\omega_o^c} \ln\left(\frac{Ph_i^c}{\omega_o^c}\right), \qquad H_2^c = \sum_{i=th+1}^{L} \frac{Ph_i^c}{\omega_1^c} \ln\left(\frac{Ph_i^c}{\omega_1^c}\right) \qquad (22)$$

$Ph_i^c$ is the probability distribution of the intensity levels which is obtained using Eq.(10). $\omega_0(th)$ and $\omega_1(th)$ are probabilities distributions for $C_1$ and $C_2$. $\ln(\cdot)$ stands for the natural logarithm. Similar to the Otsu's method the entropy-based approach can be extended for multiple threshold values, for such a case it is necessary to divide the image into $k$ classes using the similar number of thresholds. Under such conditions, the new objective function is defined as:

$$f_{Kapur}(\mathbf{TH}) = \sum_{i=1}^{k} H_i^c, \quad c = \begin{cases} 1,2,3 & \text{if} \quad \text{RGB Image} \\ 1 & \text{if} \quad \text{Gray scale Image} \end{cases} \qquad (23)$$

where $\mathbf{TH} = [th_1, th_2, ..., th_{k-1}]$ is a vector that contains the multiple thresholds. Each entropy is computed separately with its respective $th$ value, so Eq. (22) is expanded for $k$ entropies.

$$\begin{aligned} H_1^c &= \sum_{i=1}^{th_1} \frac{Ph_i^c}{\omega_o^c} \ln\left(\frac{Ph_i^c}{\omega_o^c}\right), \\ H_2^c &= \sum_{i=th_1+1}^{th_2} \frac{Ph_i^c}{\omega_1^c} \ln\left(\frac{Ph_i^c}{\omega_1^c}\right), \\ &\vdots \qquad \vdots \\ H_k^c &= \sum_{i=th_k+1}^{L} \frac{Ph_i^c}{\omega_{k-1}^c} \ln\left(\frac{Ph_i^c}{\omega_{k-1}^c}\right) \end{aligned} \qquad (24)$$

Here the values of the probability occurrence ($\omega_0^c, \omega_1^c, ..., \omega_{k-1}^c$) of the $k$ classes are obtained using Eq. (19) and the probability distribution $Ph_i^c$ with Eq. (10). Finally to separate the pixels in the respective classes it is necessary to use Eq. (9).

## 4. Multilevel Thresholding Using EMO (MTEMO)

In the proposed method, the segmentation task is faced as an optimization problem which can be stated as follows:

$$\begin{aligned} &\text{maximize} \quad f_{Otsu}(\mathbf{TH}) \text{ or } f_{Kapur}(\mathbf{TH}), \quad \mathbf{TH} = [th_1, th_2, ..., th_k] \\ &\text{subject to} \quad \mathbf{TH} \in \mathbf{X} \end{aligned} \qquad (25)$$

where $f_{Otsu}(\mathbf{TH})$ and $f_{Kapur}(\mathbf{TH})$ are the otsu (Eq.(17)) and kapur (Eq.(23) objective functions, respectively. $\mathbf{X} = \{\mathbf{TH} \in \square^k \mid 0 \leq th_i \leq 255, i=1,...,k\}$ is the bounded feasible region, constrained by the interval 0-255. Therefore, the EMO algorithm is used to find the intensity levels (**TH**) that solves the problem formulated by Eq. (X).

*4.1 Particle representation*

Each particle uses $k$ different elements, as decision variables within the optimization algorithm. Such decision variables represent a different threshold point $th$ that is used for the segmentation. Therefore, the complete population is represented as:

$$\mathbf{S}_t = [\mathbf{TH}_1^c, \mathbf{TH}_2^c, ..., \mathbf{TH}_N^c], \quad \mathbf{TH}_i^c = \left[th_1^c, th_2^c, ..., th_k^c\right]^T \qquad (26)$$

Where $t$ represents the iteration number, $T$ refers to the transpose operator, $N$ is the size of the population and $c = 1,2,3$ is set for RGB images while $c = 1$ is chosen for gray scale images. For this problem, the boundaries of the search space are set to $l = 0$ and $u = 255$, which correspond to image intensity levels.

*4.2 EMO implementation*

The proposed segmentation algorithm has been implemented considering two different objective functions, Otsu and Kapur. Therefore, the EMO algorithm has been coupled with the otsu and kapur functions, producing two different segmentation algorithms. The implementation of both algorithms can be summarized into the following steps:

**Step 1:** Read the image $I$ and if it RGB separate it into $I_R$, $I_G$ and $I_B$. If the $I$ is gray scale store it into $I_{Gr}$. $c = 1,2,3$ for RGB images or $c = 1$ for gray scale images.

**Step 2:** Obtain histograms: for RGB images $h^R$, $h^G$, $h^B$ and for gray scale images $h^{Gr}$.

**Step 3:** Calculate the probability distribution using Eq. (10) and the histograms.

**Step 4:** Initialize the EMO parameters: $Iter_{max}$, $Iter_{local}$, $\delta$, $k$ and $N$.

**Step 5:** Initialize a population $\mathbf{S}_t^c$ of $N$ random particles with $k$ dimensions.

**Step 6:** Compute the values $\omega_i^c$ and $\mu_i^c$. Evaluate $\mathbf{S}_t^c$ in the objective function $f_{Otsu}$ or $f_{Kapur}$ depending on the thresholding method.

**Step 7:** Compute the charge of each particle using Eq. (4), and with Eq. (5) and (6) compute the total force vector.

**Step 8:** Move the entire population $\mathbf{S}_t^c$ along the total force vector using Eq. (7).

**Step 9:** Apply the local search to the moved population and select the best elements of this search based on their objective function values.

**Step 10:** The $t$ index is increased in 1, If $t \geq Iter_{max}$ or if the stop criteria is satisfied the algorithm finishes the iteration process and jump to step 11. Otherwise jump to step 7.

**Step 11:** Select the particle that has the best $x_t^{B^c}$ objective function value (Eq. (3) using $f_{Otsu}$ or $f_{Kapur}$).

**Step 12:** Apply the thresholds values contained in $x_t^{B^c}$ to the image $I$ Eq. (9).

## 5. Experimental Results

The proposed algorithm has been tested under a set of 20 benchmark images. Some of these images are widely used in the image processing literature to test different methods (Lena, Cameraman, Hunter, Baboon, etc) [14, 16,]. All the images have the same size ($512 \times 512$ pixels) and they are in JPGE format.

In order to carry out the algorithm analysis the proposed MTEMO is compared to state-of-the-art thresholding methods, such Genetic Algorithms (GA) [12, 32], Particle Swarm Optimization (PSO) [2] and Bacterial Foraging (BF) [16]. Since all the methods are stochastic, it is necessary to employ an appropriate statistical metrics to compare the efficiency of the algorithms. Hence, all algorithms are executed 35 times per image, according to the related literature the number of thresholds for test are $th = 2,3,4,5$ [1, 2, and 3]. In each experiment the stop criteria is set to 50 iterations. In order to verify the stability at the end of each test the standard deviation (STD) is obtained (Eq. (27)). If the STD value increases the algorithms becomes more instable [1].

$$STD = \sqrt{\sum_{i=1}^{Iter_{max}} \frac{(\sigma_i - \mu)}{Ru}} \qquad (27)$$

On the other hand, the peak-to-signal ratio (PSNR) is used to compare the similarity of an image (image segmented) against a reference image (original image) based on the mean square error (MSE) of each pixel [3, 14, 33]. Both PSNR and MSE are defined as:

$$PSNR = 20\log_{10}\left(\frac{255}{RMSE}\right), \quad (dB)$$

$$RMSE = \sqrt{\frac{\sum_{i=1}^{ro}\sum_{j=1}^{co}\left(\mathbf{I}_o^c(i,j) - \mathbf{I}_{th}^c(i,j)\right)}{ro \times co}} \tag{28}$$

where $I_o^c$ is the original image, $I_{th}^c$ is the segmented image, $c$ depends of the image (RGB or gray scale) and $ro$, $co$ are the total number of rows and columns of the image, respectively.

The set of EMO parameters has been obtained using the criteria proposed in [17,22] and kept for all test images. Under such criteria, the parameter values are set according to a table which depends on the problem dimension. Since the maximum number of dimensions considered in this work is five, all parameters have been configured as it is shown in Table 1. According to Table 1, the minimum value of $Iter_{max}$ that guarantees the appropriate EMO operation must be 150. If such a value is incremented, it does not affect the EMO performance in terms of solution quality. Although the parameter $Iter_{max}$ represents the stop criterion since the optimization point of view, in our experiments, the stop criterion is considered as the number of times in which the best fitness values remains with no change. Therefore, if the fitness value for the best particle remains unspoiled in 10% of the total number of iterations ($Iter_{max}$), then the MTHEMO is stopped. Such a criterion has been selected to maintain compatibility to similar works reported in the literature [14-16].

| $Iter_{max}$ | $Iter_{local}$ | $\delta$ | $N$ |
|---|---|---|---|
| 150 | 10 | 0.025 | 50 |

**Table 1.** EMO Parameters.

*5.1 Otsu's results*

This section analyzes the results of MTEMO after considering the variance among classes (Eq. 17) as the objective function, just as it has been proposed by Otsu [5] ($f_{Otsu}$). The approach is applied over the complete set of benchmark images whereas the results are registered in Tables 2 and 3. Such results present the best threshold values obtained after testing the MTEMO algorithm, considering four different threshold points $th = 2,3,4,5$. The Tables 2 and 3 also features the *PSNR*, the *STD* and Iteration values. From the results, it is evident that the *PSNR* and *STD* values increment their magnitude as the number of threshold points increases. Notice that the Iteration values are the number of iterations that the algorithm needs to converge.

| Image | k | Thresholds $x_t^B$ | *PSNR* | *STD* | Iterations |
|---|---|---|---|---|---|
| Camera man | 2 | 70, 144 | 17.247 | 1.40 E-12 | 13 |
| | 3 | 58, 118, 155 | 20.226 | 3.07 E-01 | 21 |
| | 4 | 42, 95, 140, 170 | 21.533 | 8.40 E-03 | 25 |
| | 5 | 35, 82, 122, 149, 173 | 22.391 | 2.12 E+00 | 28 |
| Lena | 2 | 91, 149 | 15.480 | 0.00 E+00 | 10 |
| | 3 | 79, 125, 169 | 17.424 | 2.64 E-02 | 17 |
| | 4 | 73, 112, 144, 179 | 18.763 | 1.76 E-02 | 24 |
| | 5 | 71, 107, 135, 159, 186 | 19.442 | 6.64 E-01 | 26 |
| Baboon | 2 | 97, 149 | 15.422 | 6.92 E-13 | 15 |
| | 3 | 85, 125, 161 | 17.709 | 7.66 E-01 | 25 |
| | 4 | 71, 105, 136, 167 | 20.289 | 2.65 E-02 | 11 |
| | 5 | 66, 97, 123, 147, 173 | 21.713 | 4.86 E-02 | 22 |

| Image | k | Thresholds $x_t^B$ | PSNR | STD | Iterations |
|---|---|---|---|---|---|
| Hunter | 2 | 51, 116 | 17.875 | 2.31 E-12 | 12 |
| | 3 | 36, 86, 135 | 20.350 | 2.22 E-02 | 19 |
| | 4 | 27, 65, 104, 143 | 22.203 | 1.93 E-02 | 25 |
| | 5 | 23, 54 ,88, 112, 152 | 23.723 | 1.60 E-03 | 30 |
| Airplane | 2 | 114, 174 | 15.033 | 2.65 E-02 | 14 |
| | 3 | 92, 144, 190 | 18.854 | 9.29 E-02 | 28 |
| | 4 | 85, 130, 173, 203 | 20.717 | 1.05 E-02 | 26 |
| | 5 | 68, 106, 142, 179, 204 | 23.160 | 2.38 E-02 | 31 |
| Peppers | 2 | 72 138 | 16.299 | 1.38 E-12 | 16 |
| | 3 | 65 122 169 | 18.359 | 4.61 E-13 | 20 |
| | 4 | 50 88 128 171 | 20.737 | 4.61 E-13 | 25 |
| | 5 | 48 85 118 150 179 | 22.310 | 2.33 E-02 | 34 |
| Living Room | 2 | 87, 145 | 15.999 | 1.15 E-12 | 18 |
| | 3 | 76, 123, 163 | 18.197 | 6.92 E-12 | 24 |
| | 4 | 56, 97, 132, 168 | 20.673 | 1.78 E-01 | 29 |
| | 5 | 49, 88, 120, 147, 179 | 22.192 | 1.02 E-01 | 28 |
| Blonde | 2 | 106, 155 | 14.609 | 3.70 E-03 | 15 |
| | 3 | 53, 112, 158 | 19.157 | 9.23 E-13 | 20 |
| | 4 | 50, 103, 139, 168 | 20.964 | 2.53 E-02 | 29 |
| | 5 | 48, 95, 125, 151, 174 | 22.335 | 4.50 E-02 | 32 |
| Bridge | 2 | 91 56 | 13.943 | 4.61 E-13 | 11 |
| | 3 | 72 120 177 | 17.019 | 1.11 E+00 | 16 |
| | 4 | 63 103 145 193 | 18.872 | 3.20 E-01 | 17 |
| | 5 | 59 95 127 161 291 | 20.143 | 7.32 E-01 | 27 |
| Butterfly | 2 | 99, 151 | 13.934 | 9.68 E-02 | 10 |
| | 3 | 82, 119, 160 | 16.932 | 1.15 E-12 | 15 |
| | 4 | 81, 114, 145, 176 | 17.323 | 3.38 E+00 | 33 |
| | 5 | 61, 83, 106, 130, 163 | 21.683 | 2.86 E+00 | 25 |
| Lake | 2 | 86 155 | 14.647 | 2.53 E-02 | 18 |
| | 3 | 79 141 195 | 15.823 | 3.99 E-02 | 24 |
| | 4 | 67 111 159 199 | 17.642 | 3.91 E-02 | 32 |
| | 5 | 57 88 127 166 200 | 19.416 | 4.89 E-02 | 40 |

**Table 2.** Result after apply the MTEMO to the set of benchmark images.

| Image | k | Thresholds $x_t^B$ | PSNR | STD | Iterations |
|---|---|---|---|---|---|
| Arch monument | 2 | 70 143 | 15.685 | 2.20 E-03 | 10 |
| | 3 | 49 96 156 | 18.257 | 2.50 E-03 | 27 |
| | 4 | 42 80 126 174 | 20.190 | 1.78 E-02 | 24 |
| | 5 | 36 67 101 141 183 | 21.738 | 7.15 E-02 | 20 |
| Firemen | 2 | 61 145 | 15.511 | 9.22 E-13 | 10 |
| | 3 | 45 96 161 | 17.919 | 2.20 E-02 | 21 |
| | 4 | 43 88 139 191 | 19.832 | 1.41 E-02 | 28 |
| | 5 | 38 75 108 152 198 | 21.266 | 4.53 E-02 | 15 |
| Maize | 2 | 91 167 | 13.853 | 1.65 E-02 | 14 |
| | 3 | 76 128 187 | 15.537 | 2.16 E-02 | 18 |
| | 4 | 66 106 152 201 | 16.972 | 1.58 E-02 | 15 |
| | 5 | 58 89 126 166 209 | 18.476 | 5.75 E-02 | 53 |
| Native fisherman | 2 | 107 196 | 12.630 | 9.22 E-13 | 15 |
| | 3 | 88 135 206 | 15.015 | 6.90 E-03 | 12 |
| | 4 | 67 105 144 209 | 17.571 | 2.49 E-02 | 26 |
| | 5 | 62 96 126 157 214 | 18.835 | 2.78 E-12 | 20 |
| Pyramid | 2 | 114 167 | 12.120 | 4.61 E-13 | 16 |
| | 3 | 96 129 175 | 15.765 | 5.55 E-02 | 16 |
| | 4 | 90 119 146 186 | 17.437 | 1.98 E-02 | 40 |
| | 5 | 86 111 133 158 195 | 18.582 | 4.11 E-02 | 26 |
| Sea star | 2 | 85 157 | 14.815 | 4.61 E-13 | 15 |
| | 3 | 68 119 177 | 17.357 | 5.90 E-03 | 11 |

|  | k | Thresholds $x_i$ | PSNR | STD | Iterations |
|---|---|---|---|---|---|
|  | 4 | 60 101 138 187 | 19.125 | 5.11 E-02 | 44 |
|  | 5 | 52 86 117 150 194 | 20.729 | 5.75 E-02 | 12 |
| Smiling girl | 2 | 66 139 | 16.783 | 4.61 E-13 | 11 |
|  | 3 | 61 127 162 | 18.827 | 1.30 E-03 | 20 |
|  | 4 | 55 111 143 171 | 21.137 | 5.80 E-02 | 33 |
|  | 5 | 47 97 128 154 178 | 23.221 | 4.28 E-02 | 27 |
| Surfer | 2 | 93 163 | 12.490 | 15.7 E-02 | 22 |
|  | 3 | 71 110 176 | 15.983 | 6.92 E-13 | 21 |
|  | 4 | 47 81 118 181 | 20.677 | 2.40 E-03 | 45 |
|  | 5 | 46 77 106 143 197 | 21.864 | 5.84 E-02 | 27 |
| Train | 2 | 91 75 | 14.341 | 0.00 E+00 | 12 |
|  | 3 | 61 118 179 | 18.141 | 1.38 E-12 | 14 |
|  | 4 | 55 106 142 187 | 20.050 | 4.61 E-13 | 26 |
|  | 5 | 54 104 138 170 211 | 21.112 | 2.03 E+00 | 25 |

**Table 3.** Result after apply the MTEMO to the set of benchmark images.

For the sake of representation, it has been selected ten images of the set to show (graphically) the segmentation results. Figures 1 and 2 present the images selected from the benchmark set and their respective histograms which possess irregular distributions (see Fig. 1 (j) in particular). Under such circumstances, classical methods face great difficulties to find the best threshold values.

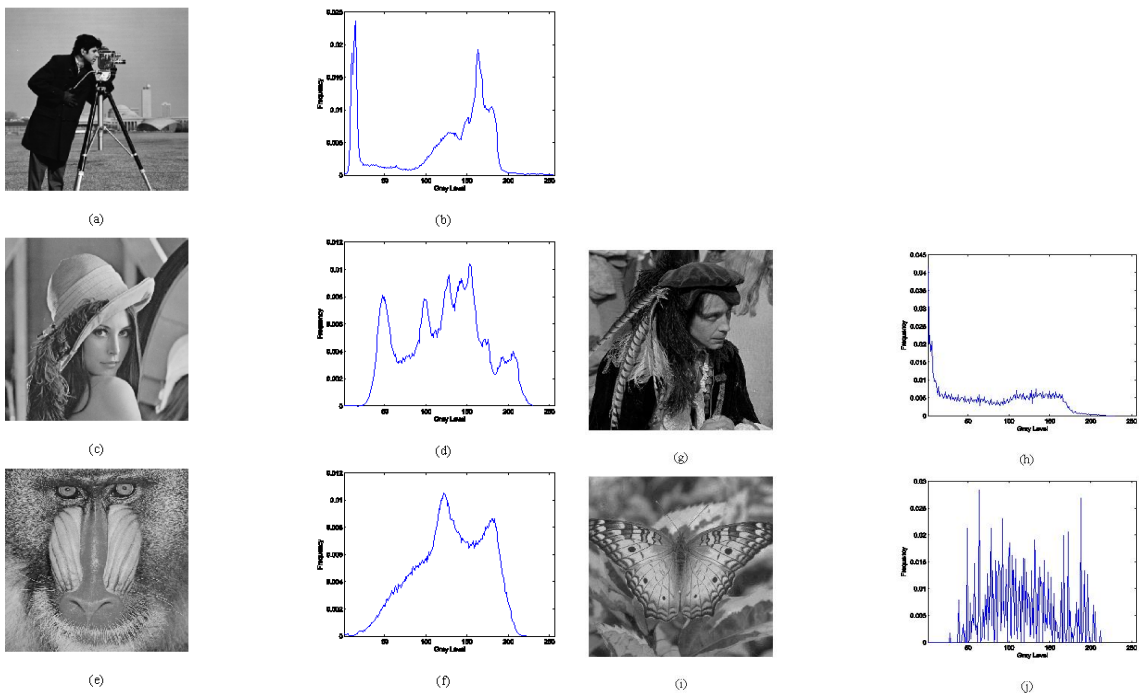

**Figure 1.** (a) Camera man, (c) Lena, (e) Baboon, (g) Hunter and (i) Butterfly, the selected benchmak images. (b), (d), (f), (h), (j) histograms of the images.

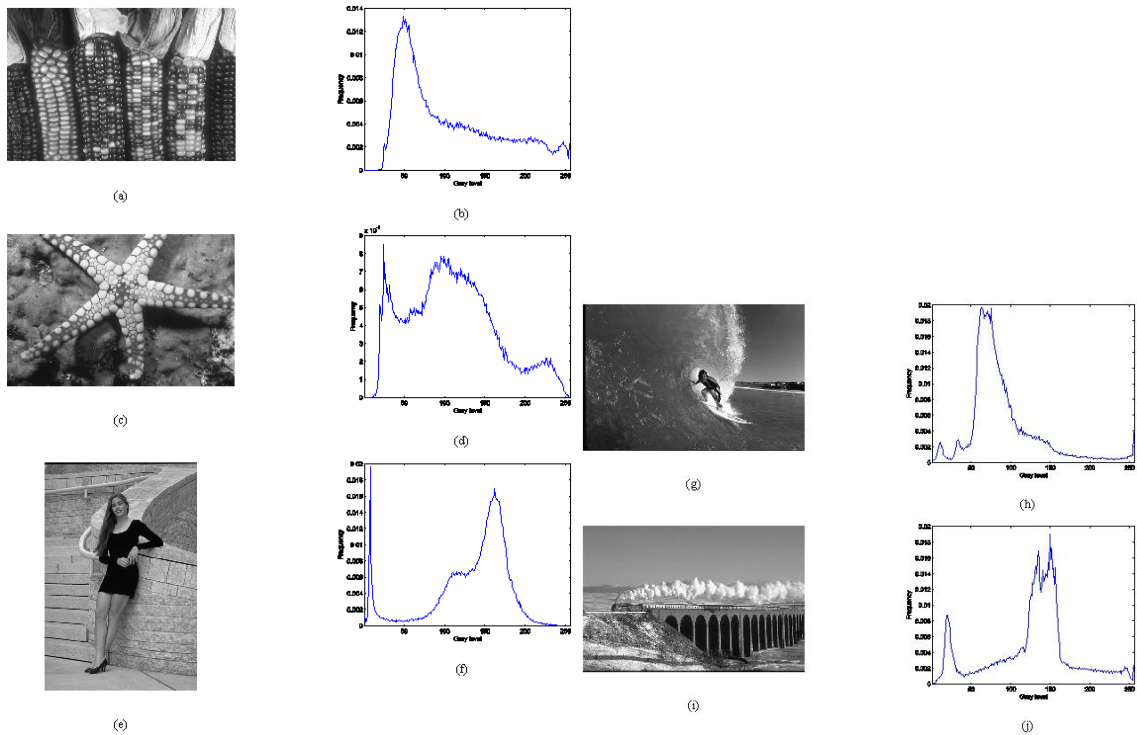

**Figure 2.** (a) Maize, (c) Sea star, (e) Smiling girl, (g) Surfer and (i) Train, the selected benchmak images. (b), (d), (f), (h), (j) histograms of the images.

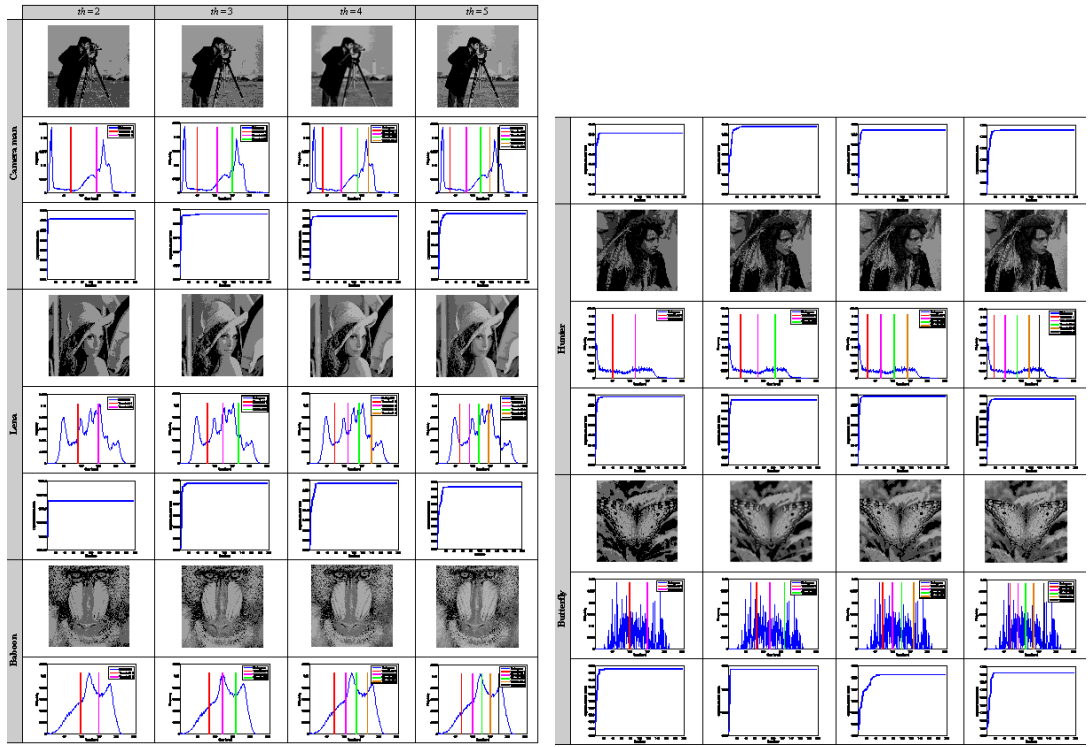

**Table 4.** Results after apply the MTEMO using Otsu´s over the selected benchamark images.

Table 4 and Table 5 show the images obtained after processing 10 original images selected from the entire benchmark set, applying the proposed algorithm. The results present the segmented images considering four different threshold points $th = 2, 3, 4, 5$. In Tables 4 and 5, it is also shown the evolution of the objective function during one execution. From the results, it is possible to appreciate that the MTEMO converges (stabilizes) around the first 50 iterations. However the algorithm continues running in order to show the convergence properties. The segmented images provide evidence that the outcome is better with $th = 4$ and $th = 5$; however, if the segmentation task does not requires to be extremely accurate then it is possible to select $th = 3$.

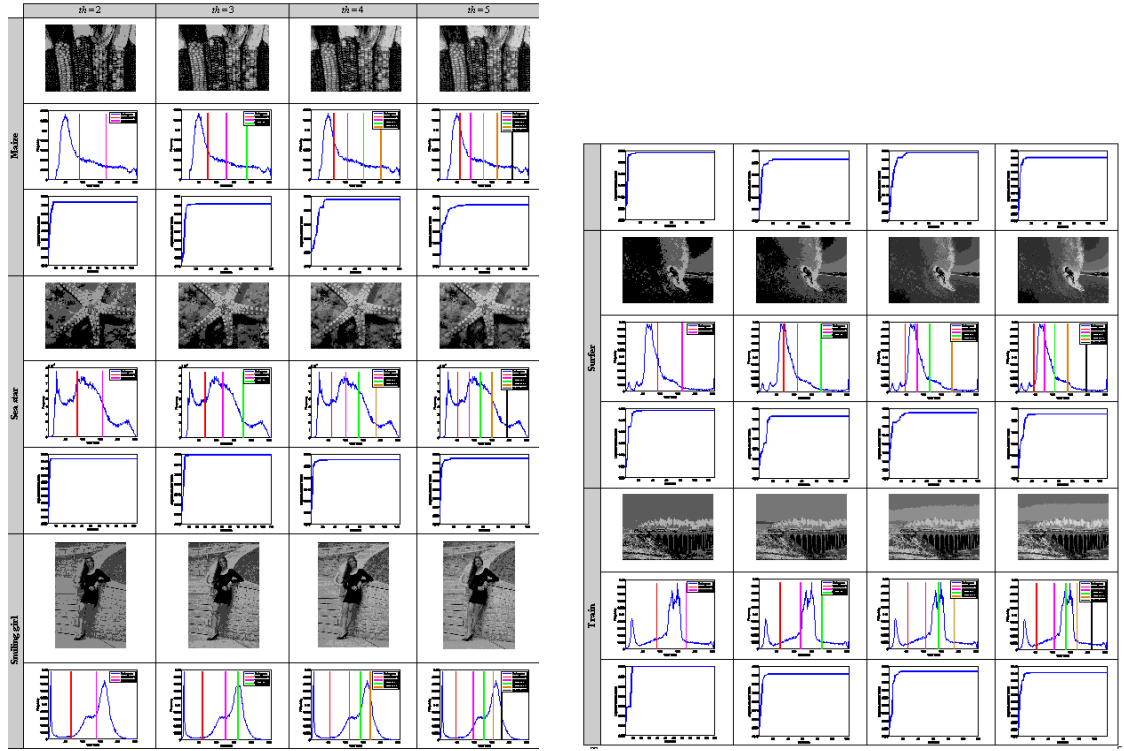

**Table 5.** Results after apply the MTEMO using Otsu´s over the selected benchamark images.

*5.2 Kapur's results*

This section analyzes the performance of MTEMO after considering as objective function (Eq. 23) the entropy function proposed by Kapur [6] ($f_{Kapur}$). In Table 6 and Table 7, are presented the experimental results after the application of MTEMO over the entire set of benchmark images. The values listed are: $PSNR$, $STD$, Iterations and the best threshold values of the last population ($x_t^B$).

| Image | k | Thresholds $x_t^B$ | *PSNR* | *STD* | Iterations |
|---|---|---|---|---|---|
| Camera man | 2 | 128, 196 | 13.626 | 3.60 E-15 | 18 |
|  | 3 | 97, 146, 196 | 18.803 | 4.91 E-02 | 25 |
|  | 4 | 44, 96, 146, 196 | 20.586 | 1.08 E-14 | 29 |
|  | 5 | 24, 60, 98, 146, 196 | 20.661 | 6.35 E-02 | 27 |
| Lena | 2 | 95, 163 | 14.672 | 0.00 E+00 | 18 |
|  | 3 | 81, 126, 176 | 17.247 | 7.50 E-04 | 25 |
|  | 4 | 76, 118, 158, 190 | 18.251 | 1.34 E-02 | 33 |
|  | 5 | 61, 92, 126, 161, 192 | 20.019 | 2.67 E-02 | 27 |
| Baboon | 2 | 79, 143 | 16.016 | 1.08 E-14 | 19 |
|  | 3 | 79, 143, 231 | 16.016 | 3.60 E-15 | 38 |
|  | 4 | 44, 98, 152, 231 | 18.485 | 2.10 E-03 | 22 |
|  | 5 | 33, 74, 114, 159, 231 | 20.507 | 1.08 E-14 | 25 |
| Hunter | 2 | 92, 179 | 15.206 | 1.44 E-14 | 17 |
|  | 3 | 59, 127, 179 | 18.500 | 4.82 E-04 | 23 |
|  | 4 | 44, 89, 133, 179 | 21.728 | 3.93 E-04 | 20 |
|  | 5 | 46, 90, 133, 179, 222 | 21.073 | 4.20 E-02 | 28 |
| Airplane | 2 | 70, 171 | 15.758 | 3,30E-03 | 18 |
|  | 3 | 68, 126, 182 | 18.810 | 1,08E-14 | 23 |

| Image | k | Thresholds $x_t^B$ | PSNR | STD | Iterations |
|---|---|---|---|---|---|
| | 4 | 68, 126, 182, 232 | 18.810 | 2,37E-01 | 30 |
| | 5 | 64, 105, 145, 185, 232 | 20.486 | 1,87E-01 | 32 |
| Peppers | 2 | 66, 143 | 16.265 | 7.21 E-15 | 15 |
| | 3 | 62, 112, 162 | 18.367 | 2.80 E-03 | 21 |
| | 4 | 62, 112, 162, 227 | 18.376 | 1.28 E-01 | 29 |
| | 5 | 48, 86, 127, 171, 227 | 20.643 | 1.37 E-01 | 32 |
| Living Room | 2 | 89 170 | 14.631 | 2.43 E-04 | 19 |
| | 3 | 47 103 175 | 17.146 | 1.08 E-10 | 25 |
| | 4 | 47 102 153 197 | 19.068 | 8.90 E-03 | 23 |
| | 5 | 42 81 115 158 197 | 21.155 | 1.00 E-02 | 28 |
| Blonde | 2 | 125, 203 | 12.244 | 1.83 E-01 | 16 |
| | 3 | 65, 134, 203 | 16.878 | 1.40 E-01 | 24 |
| | 4 | 65, 113, 155, 203 | 20.107 | 1.95 E-01 | 26 |
| | 5 | 65, 100, 134, 166, 203 | 22.138 | 1.01 E-01 | 29 |
| Bridge | 2 | 94, 171 | 13.529 | 1.05 E-02 | 18 |
| | 3 | 65, 131, 195 | 16.806 | 1.08 E-10 | 19 |
| | 4 | 53, 102, 151, 199 | 18.902 | 1.44 E-14 | 26 |
| | 5 | 36, 73, 114, 159, 203 | 20.733 | 1.75 E-03 | 24 |
| Butterfly | 2 | 120, 213 | 11.065 | 1.35 E-01 | 22 |
| | 3 | 96, 144, 213 | 14.176 | 3.56 E-01 | 29 |
| | 4 | 27, 96, 144, 213 | 16.725 | 3.45 E-01 | 36 |
| | 5 | 27, 85, 120, 152, 213 | 19.026 | 2.32 E-01 | 30 |
| Lake | 2 | 91, 163 | 14.713 | 1.44 E-14 | 19 |
| | 3 | 73, 120, 170 | 16.441 | 9.55 E-05 | 23 |
| | 4 | 69, 112, 156, 195 | 17.455 | 1.73 E-02 | 25 |
| | 5 | 62, 96, 131, 166, 198 | 18.774 | 5.45 E-02 | 36 |

**Table 6.** Result after apply the MTEMO to the set of benchmark images.

| Image | k | Thresholds $x_t^B$ | PSNR | STD | Iterations |
|---|---|---|---|---|---|
| Arch | 2 | 80 155 | 15.520 | 1.44 E-14 | 8 |
| | 3 | 64 118 174 | 17.488 | 1.80 E-14 | 29 |
| | 4 | 61 114 165 215 | 17.950 | 5.03 E-02 | 69 |
| | 5 | 48 89 130 172 217 | 20.148 | 5.93 E-02 | 17 |
| Firemen | 2 | 102 175 | 14.021 | 6.29 E-04 | 22 |
| | 3 | 72 127 184 | 17.146 | 2.90 E-02 | 10 |
| | 4 | 70 123 172 220 | 17.782 | 4.78 E-05 | 33 |
| | 5 | 53 92 131 176 221 | 20.572 | 1.44 E-14 | 40 |
| Maize | 2 | 98 176 | 13.633 | 2.26 E-04 | 25 |
| | 3 | 81 140 198 | 15.229 | 5.92 E-05 | 24 |
| | 4 | 74 120 165 211 | 16.280 | 2.97 E-04 | 38 |
| | 5 | 68 105 143 180 218 | 17.211 | 3.27 E-04 | 40 |
| Native fisherman | 2 | 68 154 | 11.669 | 7.20 E-15 | 23 |
| | 3 | 52 122 185 | 14.293 | 1.15 E-02 | 21 |
| | 4 | 48 100 150 197 | 16.254 | 1.44 E-14 | 18 |
| | 5 | 38 73 113 154 198 | 17.102 | 1.24 E-02 | 33 |
| Pyramid | 2 | 36 165 | 10.081 | 0.00 E+00 | 24 |
| | 3 | 36 110 173 | 15.843 | 0.00 E+00 | 15 |
| | 4 | 36 98 158 199 | 17.256 | 3.05 E-02 | 24 |
| | 5 | 36 88 124 161 201 | 20.724 | 5.71 E-02 | 20 |
| Sea star | 2 | 90 169 | 14.398 | 7.20 E-15 | 23 |
| | 3 | 75 130 184 | 16.987 | 1.08 E-14 | 20 |
| | 4 | 67 115 163 206 | 18.304 | 5.02 E-04 | 40 |
| | 5 | 56 94 133 172 211 | 20.165 | 7.51 E-04 | 45 |
| Smiling girl | 2 | 106 202 | 13.420 | 0.00 E+00 | 17 |
| | 3 | 94 143 202 | 18.254 | 6.06 E-05 | 22 |
| | 4 | 36 84 139 202 | 18.860 | 1.96 E-02 | 20 |
| | 5 | 36 84 134 178 211 | 19.840 | 5.42 E-02 | 22 |
| Surfer | 2 | 105 172 | 11.744 | 1.02 E-02 | 22 |
|  | 3 | 51 106 172 | 18.584 | 7.49 E-02 | 32 |
|  | 4 | 51 102 155 203 | 19.478 | 3.06 E-15 | 28 |
|  | 5 | 51 97 136 172 213 | 20.468 | 6.50 E-03 | 24 |
| Train | 2 | 105 169 | 14.947 | 0.00 E+00 | 18 |
|  | 3 | 70 120 171 | 18.212 | 8.20 E-03 | 18 |
|  | 4 | 70 120 162 208 | 19.394 | 1.44 E-14 | 26 |
|  | 5 | 39 79 121 162 208 | 20.619 | 4.56 E-02 | 24 |

**Table 7.** Result after apply the MTEMO to the set of benchmark images.

Tables 8 and 9 show the images obtained after processing 10 images selected from the entire benchmark set, applying the proposed algorithm. The results present the segmented images considering four different threshold points $th = 2, 3, 4, 5$.

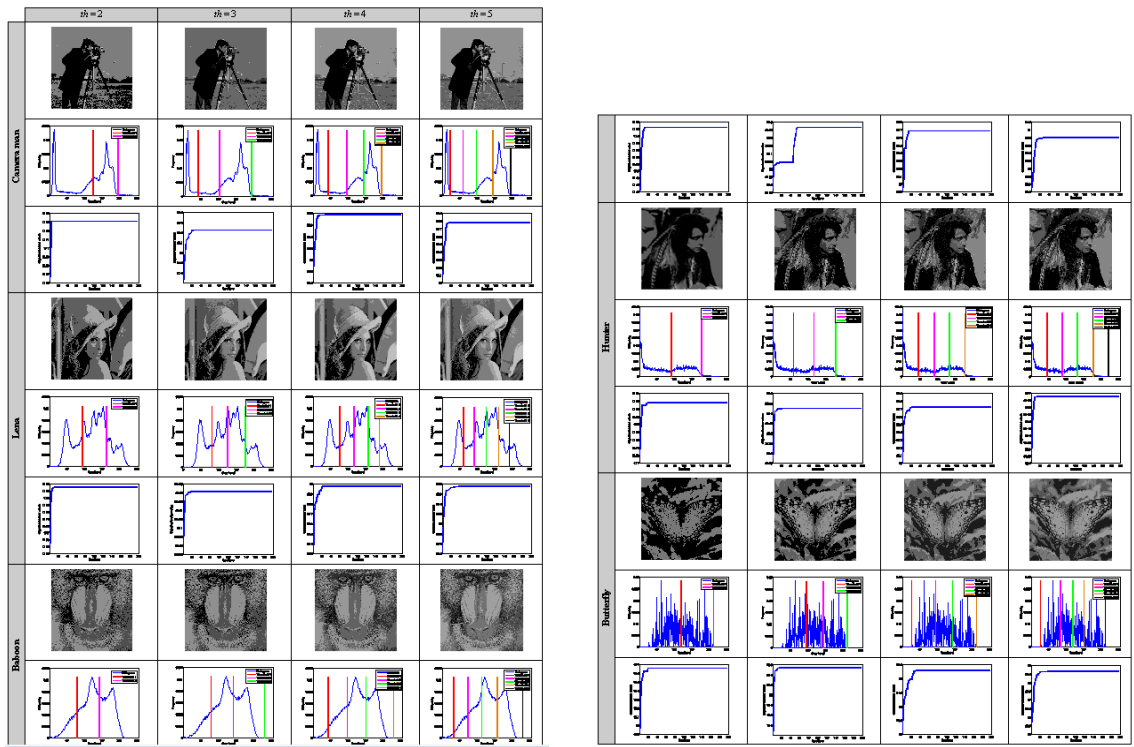

**Table 8.** Results after apply the MTEMO using Kapur´s over the selected benchamark images.

*5.4 Contaminated Images*

Another important test consist in add two different kind of noise to a selected test images. The objective is to verify if the proposed algorithm is able to segment the contaminated images. Gaussian noise is used in this test; its parameters are $\mu = 0$ (mean) and $\sigma = 0.1$ (variance). On the other hand, a 2% of Salt and Pepper (impulsive) noise is used to contaminate the selected images.

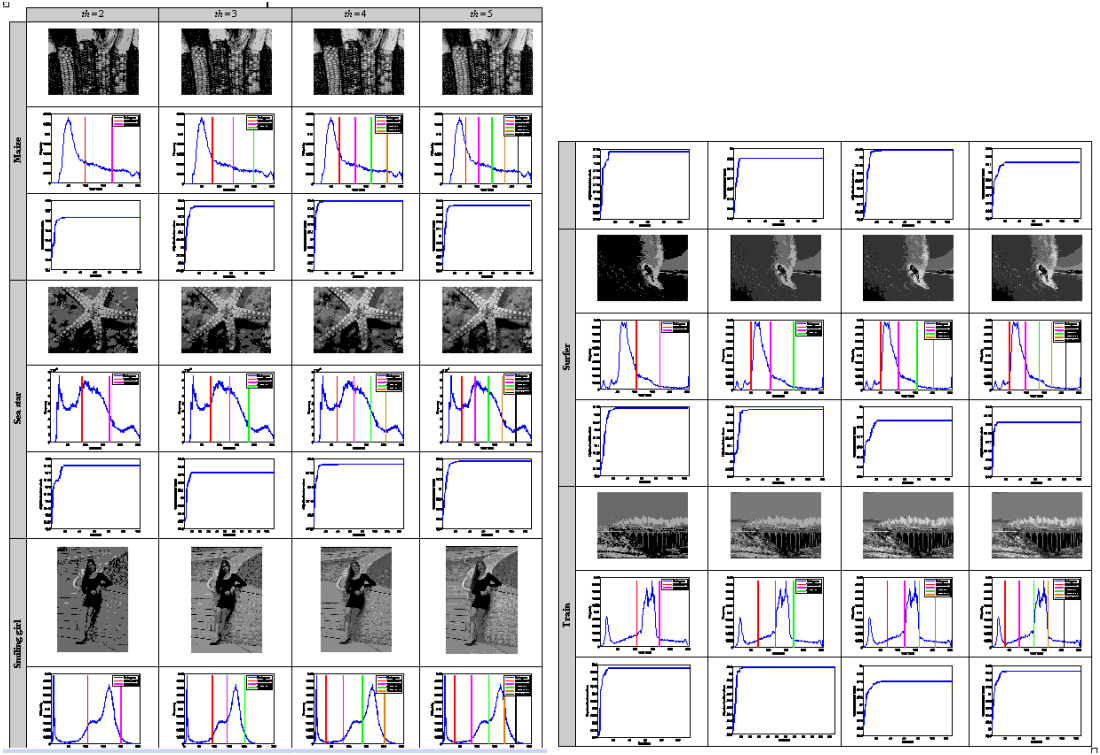

**Table 9.** Results after apply the MTEMO using Kapur´s over the selected benchamark images.

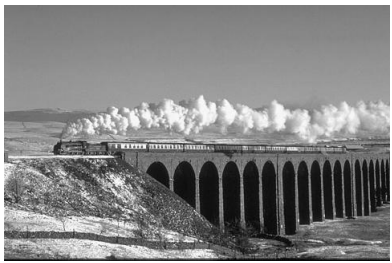

(a)

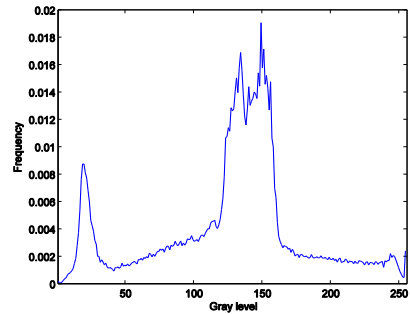

(b)

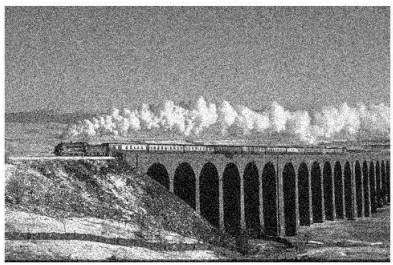

(c)

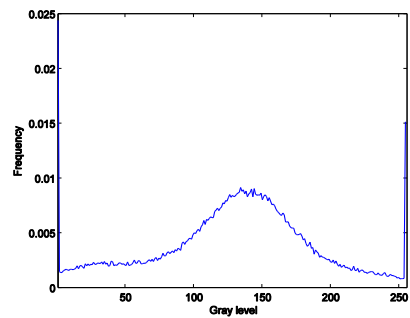

(d)

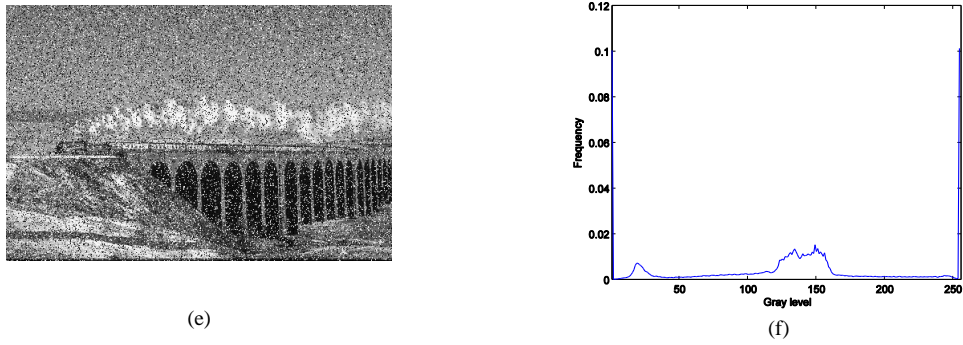

(e)  (f)

**Figure 3.** (a) Original Train image, (c) Gaussian contaminated Train image, (e) Salt and pepper contaminated Train image, (b), (d), (f) histograms of the images.

Figure 3 presents the original Train image taken from the entire benchmark set. Besides the noisy images are presented and their respective histograms. Such histograms show that they are distorted as a consequence of the added noise. Although the pixels and their distribution are modified, the results are consistent with the outcomes presented by images without noise. Table 10 shows the results after apply the proposed method with Otsu´s function over the contaminated images.

**Table 10.** Results after apply the MTEMO using Otsu´s over the noised Train image.

Table 11 presents the experimental results after using the Kapur's objective function $f_{Kapur}$ over the noised Train image, for four different $th$ values ($th = 2, 3, 4, 5$).

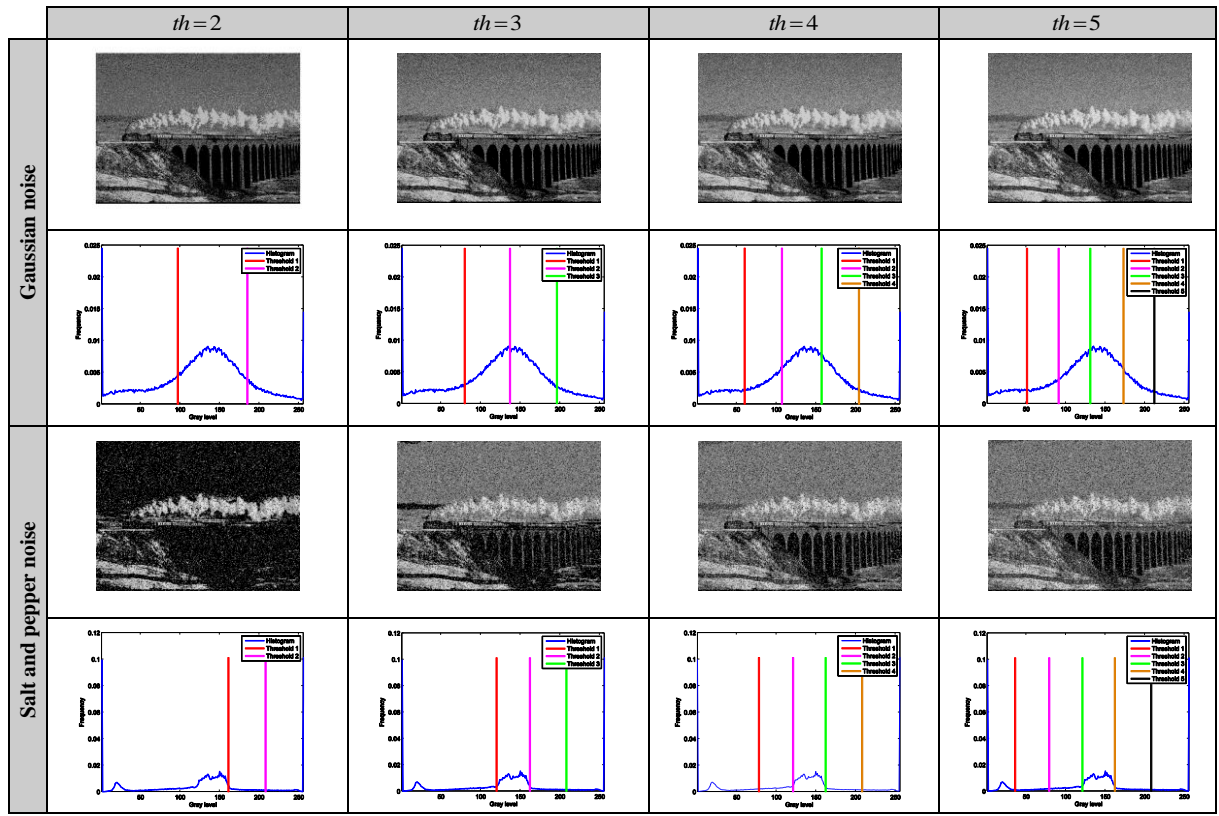

**Table 11.** Results after apply the MTEMO using Kapur´s over the noised Train image.

*5.5 Comparisons*

In order to analyse the results of the proposed approach, three different comparisons are executed. The first one involves the comparison between the two versions of MTEMO, with the Otsu function and other with the Kapur criterion. The second one analyses the comparison among the MTEMO with other state-of-the-art approaches. Finally the third one compares the number of iterations of MTEMO and the selected methods, in order to verify its performance and computational effort.

*5.5.1 Comparison between Otsu and Kapur*

In order to statistically compare the results from Tables 2, 3, 6 and 7, a non-parametric significance proof known as the Wilcoxon's rank test [34,35] for 35 independent samples has been conducted. Such proof allows assessing result differences among two related methods. The analysis is performed considering a 5% significance level over the peak-to-signal ratio (PSNR) data corresponding to the five threshold points. Table 12 reports the *p*-values produced by Wilcoxon's test for a pair-wise comparison of the PSNR values between the Otsu and Kapur objective functions. As a null hypothesis, it is assumed that there is no difference between the values of the two objective functions. The alternative hypothesis considers an existent difference between the values of both approaches. All *p*-values reported in the Table 5 are less than 0.05 (5% significance level) which is a strong evidence against the null hypothesis, indicating that the Otsu PSNR mean values for the performance are statistically better and it has not occurred by chance.

| **Image** | ***p*-Value Otsu vs. Kapur** |
|---|---|
| Camera man | 2.8061e-005 |
| Lena | 1.2111e-004 |
| Baboon | 2.6722e-004 |
| Hunter | 2.1341e-004 |
| Airplane | 8.3241e-005 |
| Peppers | 7.9341e-005 |
| Living Room | 1.4522e-004 |
| Blonde | 9.7101e-005 |
| Bridge | 1.3765e-004 |
| Butterfly | 6.2955e-005 |
| Lake | 4.7303e-005 |
| Arch | 4.9426e-005 |
| Firemen | 4.7239e-005 |
| Maize | 1.6871e-004 |
| Native fisherman | 3.5188e-004 |
| Pyramid | 9.3876e-005 |
| Sea star | 1.4764e-005 |
| Smiling girl | 7.1464e-004 |
| Surfer | 9.5993e-005 |
| Train | 3.5649e-004 |

**Table 12.** *p*-values produced by Wilcoxon's test comparing Otsu vs. Kapur over the averaged PSNR from Tables 2, 3, 6 and 7.

### 5.5.2 Comparison among MTEMO and other MT approaches

In order to demonstrate that the MTEMO is an interesting alternative for MT, the proposed algorithm is compared with other similar implementations. The other methods used in the comparison are: Genetic Algorithms (GA), Particle Swarm Optimization (PSO) and Bacterial foraging (BF).

All the algorithms run 35 times over each selected image. The images used for this test are the same of the selected in subsection 5.2 and 5.1 (Camera man, Lena, Baboon, Hunter, Butterfly, Maize, Sea star, Smiling girl, Surfer and Train). For each image is computed the *PSNR*, *STD* and the mean of the objective function values, moreover the entire test is performed using both Otsu`s and Kapur`s objective functions.

| Image | k | MTEMO | | | GA | | | PSO | | | BF | | |
|---|---|---|---|---|---|---|---|---|---|---|---|---|---|
| | | *PSNR* | *STD* | Mean | *PSNR* | *STD* | Mean | *PSNR* | *STD* | Mean | *PSNR* | *STD* | Mean |
| Camera man | 2 | **17.247** | 1.40 E-12 | 3606.3 | 17.048 | 0.0232 | 3604.5 | 17.033 | 0.0341 | 3598.3 | 17.058 | 0.0345 | 3590.9 |
| | 3 | **20.226** | 3.07 E-01 | 3679.5 | 17.573 | 0.1455 | 3678.3 | 19.219 | 0.2345 | 3662.7 | 20.035 | 0.2459 | 3657.5 |
| | 4 | **21.533** | 8.40 E-03 | 3782.4 | 20.523 | 0.2232 | 3781.5 | 21.254 | 0.3142 | 3777.4 | 21.209 | 0.4560 | 3761.4 |
| | 5 | **22.391** | 2.12 E+00 | 3767.6 | 21.369 | 0.4589 | 3766.4 | 22.095 | 0.5089 | 3741.6 | 22.237 | 0.5089 | 3789.8 |
| Lena | 2 | **15.480** | 0.00 E+00 | 1939.3 | 15.040 | 0.0049 | 1960.9 | 15.077 | 0.0033 | 1961.4 | 15.031 | 2.99 E-04 | 1961.5 |
| | 3 | **17.424** | 2.64 E-02 | 2103.8 | 17.304 | 0.1100 | 2126.4 | 17.276 | 0.0390 | 2127.7 | 17.401 | 0.0061 | 2128.0 |
| | 4 | **18.763** | 1.76 E-02 | 2166.8 | 17.920 | 0.2594 | 2173.7 | 18.305 | 0.1810 | 2180.6 | 18.507 | 0.0081 | 2189.0 |
| | 5 | **19.442** | 6.64 E-01 | 2192.4 | 18.402 | 0.3048 | 2196.2 | 18.770 | 0.2181 | 2212.5 | 19.001 | 0.0502 | 2215.6 |
| Baboon | 2 | **15.422** | 6.92 E-13 | 1548.1 | 15.304 | 0.0031 | 1547.6 | 15.088 | 0.0077 | 1547.9 | 15.353 | 8.88 E-04 | 1548.0 |
| | 3 | **17.709** | 7.66 E-01 | 1638.3 | 17.505 | 0.1750 | 1633.5 | 17.603 | 0.0816 | 1635.3 | 17.074 | 0.0287 | 1637.0 |
| | 4 | **20.289** | 2.65 E-02 | 1692.1 | 18.708 | 0.2707 | 1677.7 | 19.233 | 0.0853 | 1684.3 | 19.654 | 0.0336 | 1690.7 |
| | 5 | **21.713** | 4.86 E-02 | 1717.8 | 20.203 | 0.3048 | 1712.9 | 20.526 | 0.1899 | 1712.9 | 21.160 | 0.1065 | 1716.7 |
| Hunter | 2 | **17.875** | 2.31 E-12 | 3064.2 | 17.088 | 0.0470 | 3064.1 | 17.932 | 0.2534 | 3064.1 | 17.508 | 0.0322 | 3064.1 |
| | 3 | **20.350** | 2.22 E-02 | 3213.5 | 20.045 | 0.1930 | 3212.9 | 19.940 | 0.9727 | 3212.4 | **20.350** | 0.9627 | 3213.4 |
| | 4 | **22.203** | 1.93 E-02 | 3269.5 | 20.836 | 0.6478 | 3268.4 | 21.128 | 2.2936 | 3266.3 | 21.089 | 2.2936 | 3266.3 |
| | 5 | **23.723** | 1.60 E-03 | 3308.1 | 21.284 | 1.6202 | 3305.6 | 22.026 | 4.1811 | 3276.3 | 22.804 | 3.6102 | 3291.1 |
| Butterfly | 2 | **13.934** | 9.68 E-02 | 1553.0 | 13.007 | 0.0426 | 1553.0 | 13.092 | 0.0846 | 1553.0 | 13.890 | 0.0643 | 1553.0 |
| | 3 | 16.932 | 1.15 E-12 | 1669.3 | 15.811 | 0.3586 | 1669.0 | 17.261 | 2.6268 | 1665.7 | **17.285** | 1.2113 | 1667.2 |
| | 4 | **17.323** | 3.38 E+00 | 1709.1 | 17.104 | 0.6253 | 1709.9 | 17.005 | 3.7976 | 1702.9 | 17.128 | 2.2120 | 1707.0 |
| | 5 | **21.683** | 2.86 E+00 | 1735.0 | 18.593 | 0.5968 | 1734.4 | 18.099 | 6.0747 | 1730.7 | 18.9061 | 3.5217 | 1733.0 |
| Maize | 2 | **13.853** | 1.65 E-02 | 3562.7 | 13.014 | 0.0257 | 3500.5 | 13.693 | 6.3521 | 3560.7 | 13.712 | 0.0781 | 3459.9 |
| | 3 | **15.537** | 2.16 E-02 | 3720.2 | 15.112 | 0.1538 | 3699.7 | 15.008 | 21.504 | 3712.2 | 15.200 | 0.2789 | 3701.0 |
| | 4 | **16.972** | 1.58 E-02 | 3799.1 | 16.203 | 0.3287 | 3701.5 | 16.157 | 17.521 | 3790.9 | 16.781 | 0.3681 | 3750.8 |
| | 5 | **18.476** | 5.75 E-02 | 3843.1 | 17.953 | 0.8569 | 3799.9 | 17.740 | 14.787 | 3836.2 | 18.102 | 0.7163 | 3810.0 |
| Sea star | 2 | **14.815** | 4.61 E-13 | 2546.9 | 14.744 | 0.0879 | 2534.8 | 14.802 | 3.0898 | 2345.2 | 14.798 | 0.0091 | 2352.8 |
| | 3 | **17.357** | 5.90 E-03 | 2779.8 | 17.034 | 0.1236 | 2699.2 | 17.339 | 11.582 | 2676.3 | 17.330 | 0.0398 | 2720.8 |
| | 4 | **19.125** | 5.11 E-02 | 2865.7 | 18.482 | 0.1897 | 2820.1 | 18.112 | 19.070 | 2657.5 | 18.818 | 0.2651 | 2821.9 |
| | 5 | **20.729** | 5.75 E-02 | 2912.8 | 19.383 | 0.3647 | 2903.0 | 19.019 | 19.083 | 2890.4 | 20.760 | 1.8793 | 2895.6 |
| Smiling girl | 2 | **16.783** | 4.61 E-13 | 2107.8 | 16.248 | 0.0129 | 2103.9 | 16.701 | 0.6896 | 2067.1 | 16.548 | 0.0359 | 2105.0 |
| | 3 | **18.827** | 1.30 E-03 | 2211.5 | 18.157 | 0.2987 | 2190.8 | 18.800 | 4.4323 | 2200.2 | 18.756 | 0.1569 | 2110.3 |
| | 4 | **21.137** | 5.80 E-02 | 2264.3 | 18.816 | 0.7964 | 2250.9 | 20.323 | 11.076 | 2250.7 | 21.091 | 0.3952 | 2259.8 |
| | 5 | **23.221** | 4.28 E-02 | 2295.5 | 19.219 | 1.9871 | 2279.7 | 22.628 | 9.7178 | 2285.1 | 22.980 | 2.7816 | 2281.3 |
| Surfer | 2 | **12.490** | 15.7 E-02 | 1448.6 | 12.001 | 0.0373 | 1342.5 | 12.579 | 1.7211 | 1448.0 | 12.109 | 0.0449 | 1395.6 |
| | 3 | **15.983** | 6.92 E-13 | 1586.5 | 14.509 | 0.1782 | 1456.7 | 14.789 | 1.7653 | 1586.1 | 15.900 | 0.3890 | 1487.6 |

|  | | | | | | | | | | | | |
|---|---|---|---|---|---|---|---|---|---|---|---|---|
|  | 4 | **20.677** | 2.40 E-03 | 1665.9 | 19.987 | 0.3513 | 1569.8 | 19.965 | 14.787 | 1659.4 | 19.992 | 0.5790 | 1598.7 |
|  | 5 | **21.864** | 5.84 E-02 | 1705.9 | 20.892 | 0.4789 | 16.001 | 21.575 | 13.274 | 1699.1 | 20.980 | 1.1239 | 1690.0 |
| **Train** | 2 | **14.341** | 0.00 E+00 | 2418.0 | 13.986 | 0.0138 | 2407.5 | 13.933 | 4.1810 | 2416.6 | 14.292 | 0.0069 | 2416.9 |
|  | 3 | **18.141** | 1.38 E-12 | 2611.5 | 17.471 | 0.2715 | 2604.6 | 17.947 | 18.797 | 2606.5 | 17.992 | 0.1450 | 2610.8 |
|  | 4 | **20.050** | 4.61 E-13 | 2697.0 | 18.082 | 0.3819 | 2661.4 | 19.131 | 12.443 | 2691.9 | 19.796 | 0.7283 | 2684.2 |
|  | 5 | **21.112** | 2.03 E+00 | 2740.3 | 20.303 | 0.4418 | 2726.6 | 20.997 | 12.719 | 2732.7 | 20.778 | 0.7404 | 2727.1 |

**Table 13.** Comparisons between MTEMO, GA, PSO and BF, applied over the selected test images using Otsu`s method.

Table 13 presents the computed values for the reduced benchmark test (ten images), the values in bold represent the best values founded at the end of the entire test. It is possible to see how the MTEMO algorithm has better performance than the others. Such values are computed using the Otsu's method as a objective function. On the other hand, the same experiment has been performed using the Kapur´s method. Using the same criteria described for the Otsu´s method the algorithm runs over 35 times in each image. The results of this experiment are presented in Table 14. The results show that the proposed MTEMO algorithm is better in comparison with the GA, PSO and BF.

| | | MTEMO | | | GA | | | PSO | | | BF | | |
|---|---|---|---|---|---|---|---|---|---|---|---|---|---|
| **Image** | k | *PSNR* | *STD* | Mean | *PSNR* | *STD* | Mean | *PSNR* | *STD* | Mean | *PSNR* | *STD* | Mean |
| **Camera man** | 2 | **13.626** | 3.60 E-15 | 17.584 | 11.941 | 0.1270 | 15.341 | 12.259 | 0.1001 | 16.071 | 12.264 | 0.0041 | 16.768 |
|  | 3 | **18.803** | 4.91 E-02 | 21.976 | 14.827 | 0.2136 | 20.600 | 15.211 | 0.1107 | 21.125 | 15.250 | 0.0075 | 21.498 |
|  | 4 | **20.586** | 1.08 E-14 | 26.586 | 17.166 | 0.2857 | 24.267 | 18.000 | 0.2005 | 25.050 | 18.406 | 0.0081 | 25.093 |
|  | 5 | **20.661** | 6.35 E-02 | 30.506 | 19.795 | 0.3528 | 28.326 | 20.963 | 0.2734 | 28.365 | 21.211 | 0.0741 | 30.026 |
| **Lena** | 2 | **14.672** | 0.00 E+00 | 17.831 | 12.334 | 0.0049 | 16.122 | 12.345 | 0.0033 | 16.916 | 12.345 | 2.99 E-4 | 16.605 |
|  | 3 | **17.247** | 7.50 E-04 | 22.120 | 14.995 | 0.1100 | 20.920 | 15.133 | 0.0390 | 20.468 | 15.133 | 0.0061 | 20.812 |
|  | 4 | **18.251** | 1.34 E-02 | 25.999 | 17.089 | 0.2594 | 23.569 | 17.838 | 0.1810 | 24.449 | 17.089 | 0.0081 | 26.214 |
|  | 5 | **20.019** | 2.67 E-02 | 29.787 | 19.549 | 0.3043 | 27.213 | 20.442 | 0.2181 | 27.526 | 19.549 | 0.0502 | 28.046 |
| **Baboon** | 2 | **16.016** | 1.08 E-14 | 17.625 | 12.184 | 0.0567 | 16.425 | 12.213 | 0.0077 | 16.811 | 12.216 | 8.88 E-4 | 16.889 |
|  | 3 | **16.016** | 3.60 E-15 | 22.269 | 14.745 | 0.1580 | 21.069 | 15.008 | 0.0816 | 21.088 | 15.211 | 0.0287 | 21.630 |
|  | 4 | **18.485** | 2.10 E-03 | 26.688 | 16.935 | 0.1765 | 25.489 | 17.574 | 0.0853 | 24.375 | 17.999 | 0.0336 | 25.446 |
|  | 5 | **20.507** | 1.08 E-14 | 30.800 | 19.662 | 0.2775 | 29.601 | 20.224 | 0.1899 | 30.994 | 20.720 | 0.1065 | 30.887 |
| **Hunter** | 2 | **15.206** | 1.44 E-14 | 17.856 | 12.349 | 0.0148 | 16.150 | 12.370 | 0.0068 | 15.580 | 12.373 | 0.0033 | 16.795 |
|  | 3 | **18.500** | 4.82 E-04 | 22.525 | 14.838 | 0.1741 | 21.026 | 15.128 | 0.0936 | 20.639 | 15.553 | 0.1155 | 21.860 |
|  | 4 | **21.729** | 3.93 E-04 | 26.728 | 17.218 | 0.2192 | 25.509 | 18.040 | 0.1560 | 27.085 | 18.381 | 0.0055 | 26.230 |
|  | 5 | **21.074** | 4.20 E-02 | 30.642 | 19.563 | 0.3466 | 29.042 | 20.533 | 0.2720 | 29.013 | 21.256 | 0.0028 | 28.856 |
| **Butterfly** | 2 | **11.0653** | 1.35E-01 | 16.681 | 10.470 | 0.0872 | 15.481 | 10.474 | 0.0025 | 14.098 | 10.474 | 0.0014 | 15.784 |
|  | 3 | **14.1766** | 3.56E-01 | 21.242 | 11.628 | 0.2021 | 20.042 | 12.313 | 0.1880 | 19.340 | 12.754 | 0.0118 | 21.308 |
|  | 4 | **16.7257** | 3.45E-01 | 25.179 | 13.314 | 0.2596 | 23.980 | 14.231 | 0.2473 | 25.190 | 14.877 | 0.0166 | 25.963 |
|  | 5 | **19.0267** | 2.32E-01 | 28.611 | 15.756 | 0.3977 | 27.411 | 16.337 | 0.2821 | 27.004 | 16.828 | 0.0877 | 27.980 |
| **Maize** | 2 | **13.633** | 2.26 E-04 | 18.631 | 13.506 | 0.0725 | 18.521 | 13.466 | 0.0012 | 18.631 | 13.601 | 0.0022 | 18.625 |
|  | 3 | **15.229** | 5.92 E-05 | 23.565 | 15.150 | 0.1582 | 23.153 | 15.018 | 0.0530 | 23.259 | 15.032 | 0.0068 | 23.128 |
|  | 4 | **16.280** | 2.97 E-04 | 27.529 | 15.909 | 0.2697 | 26.798 | 15.834 | 0.1424 | 27.470 | 16.120 | 0.0128 | 27.198 |
|  | 5 | **17.211** | 3.27 E-04 | 31.535 | 16.921 | 0.8971 | 30.852 | 16.319 | 0.4980 | 31.255 | 16.985 | 0.0978 | 30.987 |
| **Sea star** | 2 | **14.398** | 7.20 E-15 | 18.754 | 14.282 | 0.0816 | 18.753 | 14.346 | 0.0002 | 18.593 | 14.280 | 0.0016 | 18.753 |
|  | 3 | **16.987** | 1.08 E-14 | 23.323 | 8.2638 | 0.1987 | 23.260 | 16.949 | 0.1723 | 23.289 | 16.319 | 0.1813 | 23.292 |
|  | 4 | **18.304** | 5.02 E-04 | 27.582 | 15.035 | 0.2691 | 26.533 | 18.389 | 0.2481 | 27.407 | 18.240 | 0.2092 | 26.938 |
|  | 5 | **20.165** | 7.51 E-04 | 31.562 | 19.005 | 0.9740 | 30.798 | 19.849 | 0.6159 | 31.288 | 19.052 | 0.3553 | 30.857 |
| **Smiling girl** | 2 | 13.420 | 0.00 E+00 | 17.334 | 13.092 | 0.0178 | 17.295 | 13.352 | 0.0368 | 17.321 | 13.370 | 0.0038 | 17.272 |
|  | 3 | **18.254** | 6.06 E-05 | 21.904 | 17.764 | 0.2179 | 21.580 | 18.201 | 0.0556 | 21.847 | 18.207 | 0.0178 | 21.847 |
|  | 4 | **18.860** | 1.96 E-02 | 26.040 | 17.923 | 0.3024 | 25.432 | 18.063 | 0.2817 | 25.815 | 18.340 | 0.2119 | 25.183 |
|  | 5 | **19.840** | 5.42 E-02 | 30.089 | 19.026 | 0.7128 | 27.940 | 19.200 | 0.5887 | 29.700 | 19.786 | 0.3813 | 28.300 |
| **Surfer** | 2 | 11.744 | 1.02 E-02 | 18.339 | 11.521 | 0.0219 | 18.237 | 11.698 | 0.1144 | 18.194 | 11.425 | 0.0489 | 18.269 |
|  | 3 | **18.584** | 7.49 E-02 | 23.231 | 17.181 | 0.1715 | 22.865 | 18.413 | 0.2332 | 22.214 | 18.509 | 0.1369 | 23.089 |
|  | 4 | **19.478** | 3.06 E-15 | 27.863 | 18.868 | 0.2093 | 26.447 | 19.125 | 0.4214 | 26.676 | 19.388 | 0.8240 | 26.859 |
|  | 5 | **20.468** | 6.50 E-03 | 31.823 | 19.521 | 0.3182 | 30.363 | 19.491 | 0.4789 | 30.587 | 19.935 | 0.9684 | 30.968 |
| **Train** | 2 | **14.947** | 0.00 E+00 | 18.574 | 14.857 | 0.0222 | 18.573 | 14.933 | 0.0004 | 18.574 | 14.795 | 0.0080 | 18.487 |
|  | 3 | **18.212** | 8.20 E-03 | 23.107 | 17.803 | 0.2084 | 22.663 | 18.185 | 0.1013 | 23.084 | 18.081 | 0.0772 | 22.009 |
|  | 4 | **19.394** | 1.44 E-14 | 27.608 | 18.932 | 0.3065 | 26.510 | 18.667 | 0.4335 | 27.335 | 19.327 | 0.2617 | 26.564 |
|  | 5 | **20.619** | 4.56 E-02 | 31.647 | 19.781 | 1.1560 | 30.196 | 20.525 | 0.4122 | 31.484 | 20.361 | 0.7846 | 30.688 |

**Table 14.** Comparisons between MTEMO, GA, PSO and BF, applied over the selected test images using Kapur`s method.

*5.5.3 Performance and computational effort among MTEMO and other MT approaches*

In this section, it is compared the performance and computational effort of the proposed method and the GA, PSO and BF approaches. Table 15 presents the required number of iterations for each algorithm to achieve a stable objective function value. In the analysis, both objective functions, Otsu's and Kapur's, are employed to find the best threshold values for each image of the complete set of test images.

| Image | k | Otsu | | | | Kapur | | | |
|---|---|---|---|---|---|---|---|---|---|
| | | MTEMO Iterations | GA Iterations | PSO Iterations | BF Iterations | MTEMO Iterations | GA Iterations | PSO Iterations | BF Iterations |
| Camera man | 2 | **13** | 184 | 132 | 90 | **18** | 237 | 93 | 131 |
| | 3 | **21** | 300 | 287 | 138 | **25** | 195 | 133 | 206 |
| | 4 | **25** | 535 | 431 | 129 | **29** | 315 | 243 | 254 |
| | 5 | **28** | 583 | 755 | 396 | **27** | 441 | 366 | 305 |
| Lena | 2 | **10** | 142 | 116 | 73 | **18** | 193 | 224 | 180 |
| | 3 | **17** | 314 | 230 | 152 | **25** | 280 | 338 | 265 |
| | 4 | **24** | 415 | 397 | 147 | **33** | 277 | 351 | 240 |
| | 5 | **26** | 620 | 386 | 335 | **27** | 476 | 422 | 308 |
| Baboon | 2 | **15** | 186 | 167 | 116 | **19** | 286 | 378 | 140 |
| | 3 | **25** | 348 | 267 | 180 | **38** | 368 | 386 | 275 |
| | 4 | **11** | 443 | 369 | 179 | **22** | 410 | 690 | 483 |
| | 5 | **22** | 632 | 518 | 288 | **25** | 789 | 755 | 518 |
| Hunter | 2 | **12** | 254 | 171 | 180 | **17** | 238 | 185 | 176 |
| | 3 | **19** | 278 | 191 | 74 | **23** | 264 | 353 | 187 |
| | 4 | **25** | 494 | 385 | 253 | **20** | 446 | 482 | 328 |
| | 5 | **30** | 803 | 406 | 356 | **28** | 659 | 884 | 364 |
| Butterfly | 2 | **10** | 240 | 173 | 112 | **22** | 290 | 300 | 217 |
| | 3 | **15** | 331 | 240 | 144 | **29** | 339 | 374 | 276 |
| | 4 | **33** | 341 | 515 | 297 | **36** | 462 | 424 | 304 |
| | 5 | **25** | 705 | 581 | 134 | **30** | 755 | 500 | 345 |
| Maize | 2 | **10** | 152 | 288 | 156 | **25** | 115 | 334 | 168 |
| | 3 | **27** | 188 | 473 | 178 | **24** | 145 | 491 | 198 |
| | 4 | **24** | 201 | 642 | 185 | **38** | 197 | 588 | 201 |
| | 5 | **20** | 225 | 921 | 235 | **40** | 208 | 811 | 195 |
| Sea star | 2 | **15** | 235 | 333 | 221 | **23** | 270 | 334 | 191 |
| | 3 | **11** | 401 | 440 | 356 | **20** | 332 | 540 | 178 |
| | 4 | **44** | 543 | 753 | 362 | **40** | 356 | 589 | 273 |
| | 5 | **12** | 606 | 703 | 470 | **45** | 496 | 828 | 315 |
| Smiling girl | 2 | **11** | 524 | 300 | 143 | **17** | 250 | 446 | 197 |
| | 3 | **20** | 472 | 549 | 269 | **22** | 340 | 681 | 341 |
| | 4 | **33** | 388 | 616 | 456 | **20** | 445 | 852 | 689 |
| | 5 | **27** | 645 | 723 | 573 | **22** | 780 | 992 | 754 |
| Surfer | 2 | **22** | 502 | 324 | 149 | **22** | 193 | 526 | 378 |
| | 3 | **21** | 431 | 535 | 193 | **32** | 235 | 622 | 493 |
| | 4 | **45** | 322 | 511 | 217 | **28** | 399 | 819 | 697 |
| | 5 | **27** | 494 | 950 | 298 | **24** | 590 | 793 | 795 |
| Train | 2 | **12** | 511 | 342 | 189 | **18** | 434 | 361 | 257 |
| | 3 | **14** | 462 | 431 | 225 | **18** | 489 | 474 | 349 |
| | 4 | **26** | 516 | 688 | 348 | **26** | 671 | 719 | 493 |
| | 5 | **25** | 599 | 794 | 458 | **24** | 719 | 951 | 544 |

**Table 15.** Iterations comparison between MTEMO, GA, PSO and BF, applied over the selected test images using Otsu´s and Kapur`s methods.

The number of iterations in Table 15 provides evidence that the MTEMO requires less iterations to find a stable value. In [17] is provided a proof that EMO requires a low number of iterations depending on the dimension of the problem. Under such circumstances, it is demonstrated that the computational cost of MTEMO is lower than GA, PSO and BF for multilevel thresholding problems. In order to statistically prove such statement, a non-parametric Wilcoxon ranking test over the number of iterations has been used. The test is divided in three groups MTEMO vs. GA, MTEMO vs. PSO and MTEMO vs. BF. The obtained *p*- values of such analysis are presented in Table 16.

| Image | k | *p*-Value MTEMO vs. GA | *p*-Value MTEMO vs. PSO | *p*-Value MTEMO vs. BF |
|---|---|---|---|---|
| Camera man | 2 | 2.8263 E-14 | 4.1495 E-12 | 1.6185 E-14 |
| | 3 | 2.5482 E-15 | 7.1815 E-11 | 3.1253 E-15 |
| | 4 | 2.0829 E-16 | 1.6967 E-14 | 1.8069 E-13 |
| | 5 | 9.2180 E-16 | 8.3666 E-16 | 2.4299 E-14 |
| Lena | 2 | 1.9023 E-16 | 6.1475 E-11 | 2.4129 E-09 |
| | 3 | 5.7370 E-15 | 8.6537 E-14 | 7.9517 E-05 |
| | 4 | 7.9129 E-14 | 6.9820E-15 | 1.7320 E-12 |
| | 5 | 3.5309 E-12 | 4.9352 E-13 | 1.9006 E-11 |
| Baboon | 2 | 3.4520 E-09 | 1.9000 E-12 | 2.0524 E-14 |
| | 3 | 9.1500 E-07 | 2.3250 E-06 | 3.6593 E-03 |
| | 4 | 6.8490 E-05 | 1.4202 E-14 | 9.5561 E-11 |
| | 5 | 3.6003 E-08 | 1.1213 E-14 | 9.9423 E-14 |
| Hunter | 2 | 6.1892 E-13 | 3.9321 E-16 | 8.1806 E-06 |
| | 3 | 4.4766 E-13 | 6.7790 E-15 | 5.4107 E-09 |
| | 4 | 7.4115 E-14 | 7.0460 E-13 | 5.2770 E-14 |
| | 5 | 8.3869 E-15 | 7.6724 E-15 | 5.6934 E-13 |
| Butterfly | 2 | 1.4179 E-15 | 8.4310 E-09 | 7.5611 E-12 |
| | 3 | 3.0199 E-08 | 1.2170 E-04 | 9.6050 E-08 |
| | 4 | 3.7441 E-11 | 5.0935 E-12 | 8.2234 E-13 |
| | 5 | 5.1381 E-08 | 7.3796 E-15 | 4.8668 E-09 |
| Maize | 2 | 7.9676E-11 | 7.1349 E-16 | 3.2984 E-08 |
| | 3 | 9.0006 E-11 | 2.9541 E-06 | 4.6093 E-11 |
| | 4 | 9.0030 E-07 | 6.9312 E-04 | 6.8892 E-15 |
| | 5 | 1.5321 E-14 | 9.3836 E-13 | 8.2699 E-04 |
| Sea star | 2 | 1.8347 E-15 | 9.2729 E-15 | 9.6341 E-06 |
| | 3 | 2.1182 E-13 | 1.1408 E-12 | 9.6717 E-16 |
| | 4 | 3.2643 E-07 | 2.5590 E-14 | 3.9884 E-16 |
| | 5 | 7.6816 E-16 | 8.6944 E-12 | 6.4834 E-04 |
| Smiling girl | 2 | 3.1091 E-14 | 9.1850 E-08 | 7.9916 E-06 |
| | 3 | 3.3765 E-16 | 3.8180 E-06 | 8,8123 E-08 |
| | 4 | 7.3174 E-11 | 6.2570 E-07 | 4.1653 E-14 |
| | 5 | 8.5530 E-09 | 7.9818 E-08 | 2.7146 E-12 |
| Surfer | 2 | 3.4667 E-08 | 5.0517 E-16 | 9.7685 E-13 |
| | 3 | 7.3319 E-14 | 1.1479 E-13 | 2.5258 E-15 |
| | 4 | 8.8110 E-13 | 3.1081 E-14 | 3.3225 E-15 |
| | 5 | 2.6798 E-11 | 6.4653 E-09 | 3.5506 E-17 |
| Train | 2 | 3.0442 E-13 | 7.9150 E-17 | 7.4060 E-09 |
| | 3 | 4.6265 E-12 | 8.2253 E-16 | 7.6292 E-12 |
| | 4 | 5.5065 E-12 | 6.8620 E-17 | 1.6333 E-09 |
| | 5 | 8.1792 E-07 | 9.5124 E-13 | 4.6672 E-07 |

**Table 16.** *p*-values produced by Wilcoxon's test comparing Otsu vs. Kapur over the averaged PSNR from Tables 2, 3, 6 and 7.

Since the results reported on Table 16 are less than 0.05 (5% significance level), they show a strong evidence against the null hypothesis. This indicates that the number of iterations spend by MTEMO are statistically lower than its counterparts.

## 6. Conclusions

In this paper, a multilevel thresholding (MT) method based on the Electro-magnetism-Like algorithm (EMO) is presented. The approach combines the good search capabilities of EMO algorithm with the use of some objective functions that have been proposed by the popular MT methods of Otsu and Kapur. In order to measure the performance of the proposed approach, it is used the peak signal-to-noise ratio (PSNR) which assesses the segmentation quality, considering the coincidences between the segmented and the original images.

The study explores the comparison between the two versions of MTEMO, one using the Otsu objective function and the other with the Kapur criterion. The results show that the Otsu function presents better results than the Kapur criterion. Such conclusion was statistically proved considering the Wilconxon test.

The proposed approach has been compared to other techniques that implement different optimization algorithms like GA, PSO and BF. The efficiency of the algorithms was evaluated in terms of the PSNR and the STD values. The experimental results provide evidence on the outstanding performance, accuracy and convergence of the proposed algorithm in comparison to other methods. On the other hand, is proved that the computational cost of MTEMO is lower than other evolutionary approaches used in the comparison. Although the results offer evidence to demonstrate that the EMO method can yield good results on complicated images, the aim of our paper is not to devise a multilevel thresholding algorithm

Please cite this article as:
**Diego Olivaa, Erik Cuevas, Gonzalo Pajares, Daniel Zaldivar, Valentín Osuna. A Multilevel Thresholding algorithm using electromagnetism optimization**, *Neurocomputing*, **139**, (2014), 357-381.that could beat all currently available methods, but to show that electro-magnetism systems can be effectively considered as an attractive alternative for this purpose.

**Acknowledgments**

The first author acknowledges The National Council of Science and Technology of Mexico (CONACyT) for partially support this research under the doctoral grant number: 215517.**References**

[1] P. Ghamisi , M. S. Couceiro, J. A. Benediktsson, N. M.F. Ferreira, An efficient method for segmentation of images based on fractional calculus and natural selection, Expert Systems with Applications 39 (2012) 12407-12417.

[2] K. Hammouche, M. Diaf, P. Siarry, A comparative study of various meta-heuristic techniques applied to the multilevel thresholding problem, Engineering Applications of Artificial Intelligence 23 (2010) 676-688.

[3] B. Akay, A study on particle swarm optimization and artificial bee colony algorithms for multilevel thresholding, Applied Soft Computing (2012), doi 10.1016/j.asoc.2012.03.072.

[4] P-S. Liao, P-C. Chung, T-S. Chen, A fast algorithm for multilevel thresholding, Journal of Information Science and Engineering 17 (2001) 713–727.

[5] N. Otsu. A threshold selection method from gray-level histograms. IEEE Transactions on Systems, Man, Cybernetics (1979), SMC-9, 62–66.

[6] J. N. Kapur, P. K. Sahoo, A. K. C. Wong, A. K. C.  A new method for gray-level picture thresholding using the entropy of the histogram. Computer Vision Graphics Image Processing, 2 (1985) 273–285.

[7] J. Kittler & J. Illingworth. Minimum error thresholding. Pattern Recognition 19 (1986) 41–47.

[8] M, Sezgin and B. Sankur. Survey over image thresholding techniques and quantitative performance evaluation. Journal of  Electronic Imaging, 13 (2004) 146–168.

[9] P.D. Sathya, R. Kayalvizhi, Optimal multilevel thresholding using bacterial foraging algorithm, Expert Systems with Applications. Volume 38 (2011) 15549-15564.

[10] C. Lai, D. Tseng. A Hybrid Approach Using Gaussian Smoothing and Genetic Algorithm for Multilevel Thresholding. International Journal of Hybrid Intelligent Systems. Volume 1 (2004) 143-152.

[11] D. E. Goldberg. Genetic Algorithms in Search, Optimization and Machine Learning (1st ed.). Addison-Wesley Longman Publishing Co (1989), Inc., Boston, MA, USA.

[12] Peng-Yeng Yin. A fast scheme for optimal thresholding using genetic algorithms. Signal Processing, Volume 72 (1999) 85-95.

[13] J .Kennedy and R. C. Eberhart. Particle swarm optimization. Proceedings of IEEE International Conference on Neural Networks, Piscataway (1995) 1942-1948.

[14] M. Horng. Multilevel thresholding selection based on the artificial bee colony algorithm for image segmentation. Expert Systems with Applications, Volume 38 (2011) 13785-13791.

[15] S. Das, A. Biswas, S. Dasgupta, A. Abraham. Bacterial Foraging Optimization Algorithm: Theoretical Foundations, Analysis, and Applications. Studies in Computational Intelligence, Volume 203 (2009) 23-55.

[16] P. D. Sathya, R. Kayalvizhi. Optimal multilevel thresholding using bacterial foraging algorithm. Expert Systems with Applications. Volume 38 (2011) 15549-15564.24